\documentclass[journal]{IEEEtran}

\newcommand{\ELBO}{\mathrm{ELBO}}
\newcommand{\BCE}{\mathrm{BCE}}
\newcommand{\MSE}{\mathrm{MSE}}
\newcommand{\enc}{\mathrm{enc}}
\newcommand{\dec}{\mathrm{dec}}

\newcommand{\T}{\top} %
\newcommand{\orcid}[1]{\href{https://orcid.org/#1}{\includegraphics[width=0.6em]{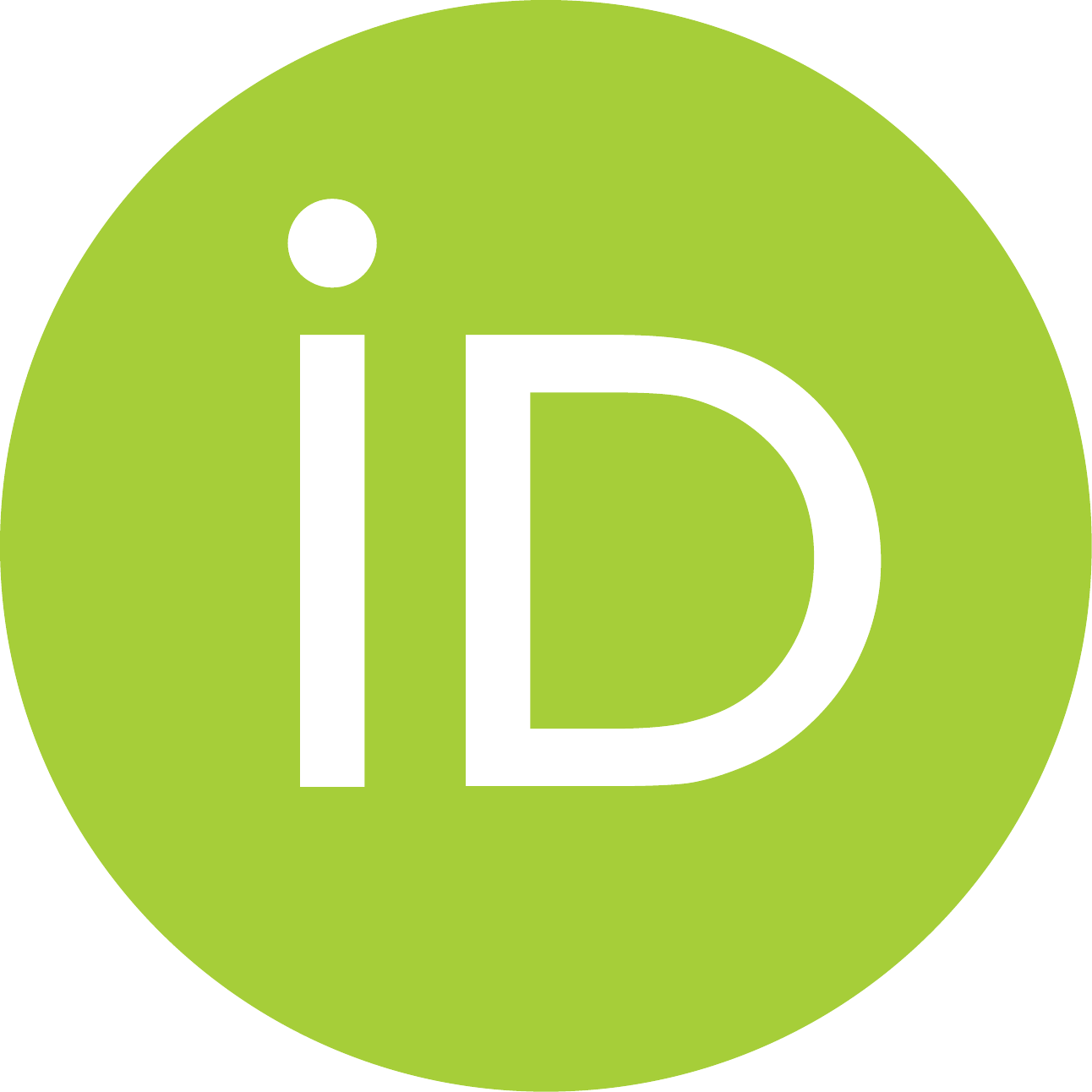}}}

\usepackage[utf8]{inputenc}

\usepackage{cite}  %
\usepackage{graphics}
\usepackage{epsfig}
\usepackage{times}
\usepackage{amsmath}
\usepackage{amssymb}
\usepackage{subcaption}
\usepackage[bookmarks=false]{hyperref}
\usepackage{siunitx}
\usepackage[ruled,vlined]{algorithm2e}
\usepackage{copyright}
\usepackage{flushend} %

\hypersetup
{
	pdftitle = {VAE-Loco: Versatile Quadruped Locomotion by Learning a Disentangled Gait Representation},
	pdfauthor = {Alexander L. Mitchell},
	pdfkeywords = {},
	pdftoolbar = true,
	colorlinks = true,
	linkcolor = black,
	citecolor = black,
	urlcolor = black,
}

\usepackage{amsmath,amsfonts,bm}

\def\eqref#1{equation~\ref{#1}}

\def\1{\bm{1}}

\def\rva{{\mathbf{a}}}

\def\rvc{{\mathbf{c}}}

\def\rvq{{\mathbf{q}}}

\def\rvs{{\mathbf{s}}}

\def\rvx{{\mathbf{x}}}

\def\rvz{{\mathbf{z}}}

\def\rmC{{\mathbf{C}}}

\def\rmI{{\mathbf{I}}}

\def\rmQ{{\mathbf{Q}}}

\def\rmS{{\mathbf{S}}}

\def\rmX{{\mathbf{X}}}

\DeclareMathAlphabet{\mathsfit}{\encodingdefault}{\sfdefault}{m}{sl}
\SetMathAlphabet{\mathsfit}{bold}{\encodingdefault}{\sfdefault}{bx}{n}

\newcommand{\KL}{D_{\mathrm{KL}}}

\newcommand{\revision}[1]{{\color{black}#1}}
\newcommand{\rev}[1]{{\color{black}#1}}

\begin{document}

\title{
    VAE-Loco: Versatile Quadruped Locomotion by Learning a Disentangled Gait Representation
}

\author{
    Alexander L. Mitchell\orcid{0000-0002-8716-4598}\quad
    Wolfgang Merkt\orcid{0000-0003-3235-4906}\quad
    Mathieu Geisert\orcid{0000-0002-5651-8736}\quad
    Siddhant Gangapurwala\orcid{0000-0002-1308-3744}%
    \\
    Martin Engelcke\orcid{0000-0001-8306-1236}\quad
    Oiwi Parker Jones\orcid{0000-0003-0307-9837}\quad
    Ioannis Havoutis\orcid{0000-0002-4371-4623}\quad
    Ingmar Posner\orcid{0000-0001-6270-700X}%
    \thanks{%
    This work was supported by a UKRI/EPSRC Programme Grant [EP/V000748/1], the EPSRC grant `Robust Legged Locomotion' [EP/S002383/1], the EPSRC CDT [EP/L015897/1], the UKRI/EPSRC RAIN [EP/R026084/1] and ORCA [EP/R026173/1] Hubs and the EU H2020 Project MEMMO (780684). It was conducted as part of ANYmal Research, a community to advance legged robotics.
    \textit{(Corresponding author: Alexander L. Mitchell.)}
    }
    \thanks{
    Alexander L. Mitchell, Wolfgang Merkt, Siddhant Gangapurwala, Oiwi Parker Jones, Ioannis Havoutis, and Ingmar Posner are with the Oxford Robotics Institute, Department of Engineering Science, University of Oxford, U.K. (e-mail: \href{mailto:mitch@robots.ox.ac.uk}{mitch@robots.ox.ac.uk};
    \href{mailto:wolfgang@robots.ox.ac.uk}{wolfgang@robots.ox.ac.uk}; \href{mailto:siddhant@robots.ox.ac.uk}{siddhant@robots.ox.ac.uk}, \href{mailto:oiwi@robots.ox.ac.uk}{oiwi@robots.ox.ac.uk}, \href{mailto:ioannis@robots.ox.ac.uk}{ioannis@robots.ox.ac.uk}, \href{mailto:ingmar@robots.ox.ac.uk}{ingmar@robots.ox.ac.uk}).
    }
    \thanks{
    Mathieu Geisert is with Agility Robotics, U.S.A. Work done while at Oxford.
    }
    \thanks{
    Martin Engelcke is with DeepMind Technologies Ltd., London, U.K. Work done while at Oxford.
    }
}

\maketitle

\begin{abstract}
Quadruped locomotion is rapidly maturing to a degree where robots \rev{are able to realise highly dynamic manoeuvres}. However, current planners are unable to vary \rev{key gait parameters of the in-swing feet \emph{midair}}. In this work we address this limitation \rev{and show that it is pivotal in increasing controller robustness} by learning a latent space capturing the key stance phases constituting a particular gait. This is achieved via a generative model trained on a single trot style, which encourages disentanglement such that application of a drive signal to a single dimension of the latent state induces holistic plans synthesising a continuous variety of trot styles. We demonstrate that specific properties of the drive signal map directly to gait parameters such as cadence, footstep height and full stance duration. Due to the nature of our approach these synthesised gaits are continuously variable online during robot operation. The use of a generative model facilitates the detection and mitigation of disturbances to provide a versatile and robust planning framework. We evaluate our approach on two versions of the real ANYmal quadruped robots and demonstrate that our method achieves a continuous blend of dynamic trot styles whilst being robust and reactive to external perturbations.
\end{abstract}

\IEEEpeerreviewmaketitle
\copyrightnotice

\section{Introduction}
\begin{figure}[tb]
    \centering
    \includegraphics[width=1.0\linewidth]{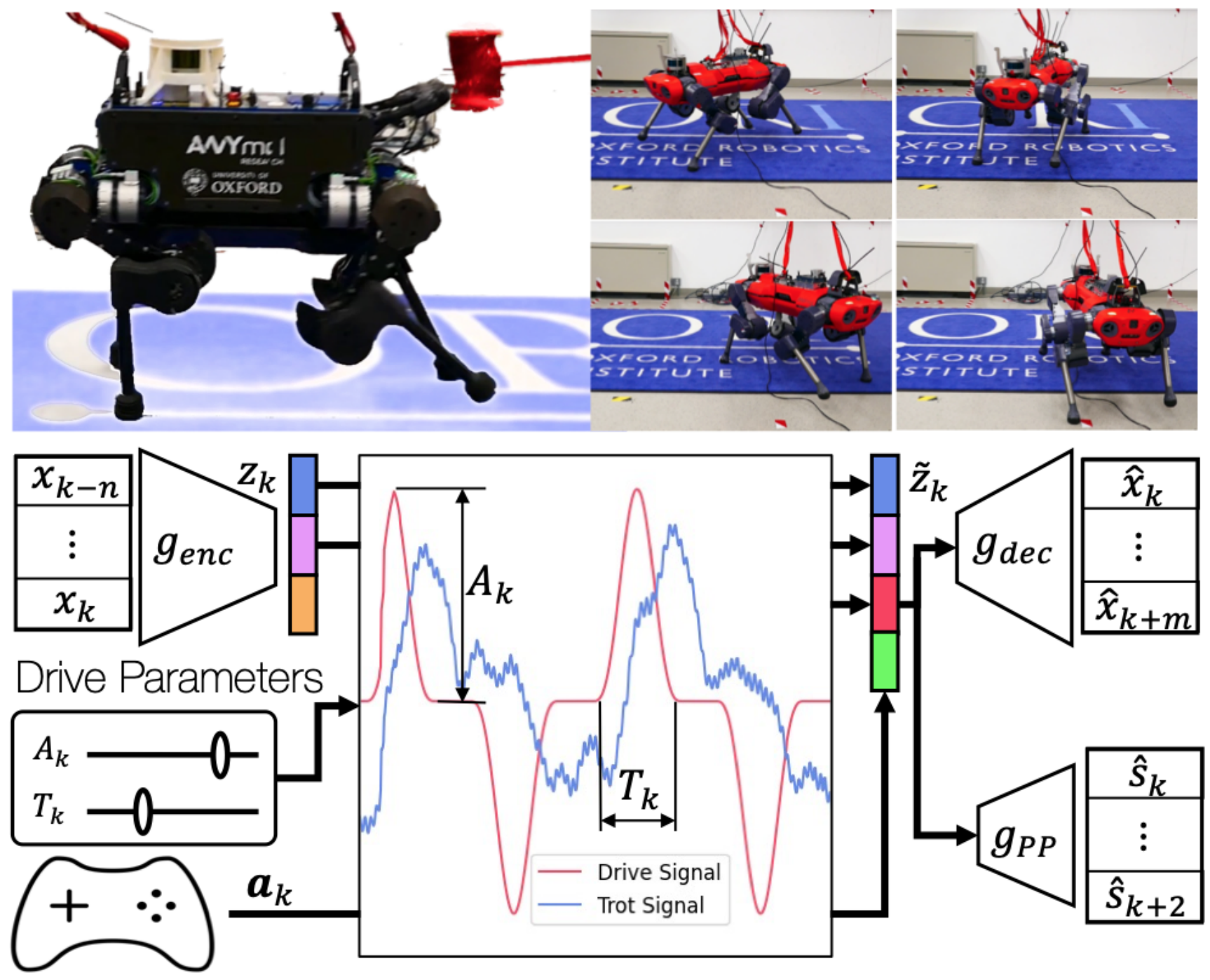}
    \caption{
    Using a variational auto-encoder (VAE), our approach learns a structured latent space capturing key stance phases constituting a particular gait. The space is disentangled to a degree such that application of a \emph{drive signal} to a single dimension of the latent variable induces gait styles which can be seamlessly interpolated between \revision{with continuous control over key gait parameters such as cadence, full-support duration, foot swing height, and footstep distance}. 
    This approach \revision{allows for precise base twist control and} readily transfers from ANYmal B to ANYmal C, a dynamically dissimilar robot, without retraining.
    Additionally, we measure disturbances as out of distribution seen during training and adjust cadence as a rudimentary, but effective response.
    }
    \label{fig:first_figure}
    \vspace{-0.5cm}
\end{figure}

Quadruped locomotion has advanced significantly in recent years, extending their capability towards applications of significant value to industry and the public domain. Driven primarily by advances in optimisation-based~\cite{bellicoso2018dynamic,boxFDDP,melon2021receding,towr,mastalli2022agile} and reinforcement learning-based methods~\cite{hwangbo2019learning,gcpo,gangapurwala2020rloc}, quadrupeds are now able to robustly plan \rev{and perform dynamic manoeuvres, making} them an increasingly popular \rev{and reliable} choice for tasks such as inspection, monitoring, search and rescue or goods delivery in difficult. However, despite recent advances, important limitations remain. Due to the complexity of the system, models used for gait planning and control are often overly simplified and handcrafted for particular \revision{and pre-determined contact schedules}~\cite{bellicoso2018dynamic,fastTraj}. In the worst case, this can limit the versatility of the robot as the models deployed are failing to exploit the full capability of the underlying hardware (e.g.~\cite{bellicoso2018dynamic,ZMP,centroidal_dynamics}). \revision{In particular, fixed gait parameters limit the ability of the robot to react to and reject external disturbances such as pushes to the robot's base. For example, the ability to adjust the robot's swing trajectories \emph{on demand} (e.g. the swing height, length and timing) allows the robot to stabilise itself. The feet can be placed faster and into positions ahead of the centre of mass. There are works which are capable of making adjustments to a pre-determined contact schedule. The first category of method updates the contact schedule utilising simplified dynamics-models~\cite{rathod2021Model}. The second type solves for the contact timings in advance by adjusting the switching times between optimisation segments~\cite{Farshidian2017An,Mastalli2017Trajectory} and uses reduced-order models. The last category optimises dynamics over gait schedules, footstep lengths and heights. These are often computationally expensive~\cite{towr,melon2021receding} meaning that varying the gait parameters is not achievable in real time.
A limitation of all these methods is that they are unable to adjust key gait parameters, in particular the contact timings, of the feet \emph{midair}. Furthermore, these adjustments are limited to small perturbations around a default value. The result of which is that these methods are unable to react quickly to external perturbations \rev{irrespective of the terrain the robot is traversing}.}
In contrast, the ability to control key gait parameters -- such as contact timing, swing height, and full-support duration -- \emph{on-the-fly} would enable a smooth interpolation between dynamic manoeuvres allowing for swift reaction to external stimuli. This leads to significantly more versatile locomotion.

Inspired by recent work on a quadruped that achieves a crawl gait via the traversal of a learnt latent space~\cite{first-steps}, we approach the challenge of continuous contact schedule variation from the perspective of learning and traversing a structured latent-space. This is enabled by learning a \emph{generative model} of locomotion data which, in addition to capturing relevant structure in the latent space, also enables the detection and mitigation of disturbances to provide a versatile and robust planning framework. In particular, we train a variational auto-encoder (VAE)~\cite{vae,vae_1} on short sequences of state-space trajectories taken from a single gait type (trot), and predict a set of future states. We show that the resulting latent space has an interpretable structure, lending itself to the generation of a variety of trot styles, depending on how the latent space is traversed. In fact, examining trot trajectories in latent space reveals an oscillatory drive signal which controls fundamental aspects of the gait. We subsequently find that by overwriting this trajectory with a synthetic drive signal, we can \emph{continuously} control the robot's gait properties whilst the robot is executing the motion. Parameters of this drive signal can be mapped explicitly to gait parameters such as cadence, footstep height, and full-stance support duration. 
\revision{With this, we can continuously vary the contact timings of feet midair. In fact we can accelerate the cadence of the swing feet from \emph{four} steps a second at take off to \emph{eight} at touch down within the duration of a single footstep. This constitutes a novel capability in quadruped control.}
We emphasise that this ability to generalise over gait styles emerges from training on a single gait type: a trot gait with constant parameters.

We illustrate the efficacy of our approach by generating a range of continuously blended trajectories firstly on the real ANYmal B quadruped robot -- a medium-sized platform (\SI{35}{\kilo \gram}) standing \SI{0.5}{\meter} tall. 
Subsequently with no retraining, we repeat the experiment on the heavier ANYmal C quadruped, which weighs in at \SI{50}{\kilo \gram} and delivers twice the peak torque of the former platform.
This demonstrates not only transfer from simulation to the real robot, but to a dynamically dissimilar platform crucially without retraining.
While the latent space is learnt using examples only from a specific gait style, our approach is able to synthesise behaviours significantly beyond this training distribution. 
\rev{We choose to limit our analysis to locomotion on flat ground in order to fully explore and understand this novel control paradigm.}

In addition, we leverage our generative approach to both characterise and react to external perturbations. 
A large impulse applied to the robot's base triggers a spike in the Evidence Lower Bound (ELBO) which clearly identifies the disturbance as out of the distribution seen during training.
Inspired by \cite{Moyer2006GaitPA}, which states that an increase in cadence is both a response to slip and a form of push recovery in humans, our planner automatically increases the robot's cadence to aid in counteracting the disturbance. This demonstrates a marked improvement in robustness.

To the best of our knowledge, our method is the first which \revision{enables the continuous online adaptation} of the robot's gait characteristics \revision{during a swing phase} whilst the robot is walking. It provides a versatile and data-driven approach to quadruped locomotion which additionally allows for disturbance detection and recovery. 

\subsection{Statement of Contributions}
\rev{The primary contribution of this paper and its conference version \emph{Next Steps}~\cite{NextSteps} is a novel planning methodology, which realises locomotion with continuously variable gait parameters.
Indeed, prior art is only able to alter the future timings of quadruped’s contact schedule~\cite{rathod2021Model,Farshidian2017An}, whilst the methods presented here facilitate the variation of timings for in-swing feet midair. Varying the foot step speed of the airborne feet in response to detected disturbance is shown to increase the robot’s ability to reject large push disturbances.}
\rev{In this paper, we deploy the VAE without retraining} on a brand new platform with significantly different dynamics: the ANYmal C.

This paper extends our previous work, \emph{Next Steps}~\cite{NextSteps} in order to showcase in-depth analysis of the method previously proposed. This includes significant extensions to key sections of Next Steps as well as new analysis.
In particular, we have extended the description and justification for the oscillatory drive-signal in Sec.~\ref{section:drive-signal}. 
We have significantly expanded the interpretation of the latent-space structure and explain how the trajectory in this space looks (Sec.~\ref{section:exp_latent_space_structure}).
Additionally, we compare the nominal trajectory in latent space to that seen during the disturbance and recovery phase, significantly expanding on the analysis of the disturbance rejection experiments (Sec.~\ref{section:disturbance_rejetion}).
We also extend the discussion of the ablation study of the model's sensitivity to hyper-parameters (Sec.~\ref{section:ablation_study}).

In this paper, we introduce new aspects of analysis. 
Firstly, we \revision{show} how backpropagating the binary cross-entropy (BCE) through the VAE's encoder \revision{is essential for structuring} the latent-space (Sec.~\ref{section:bce_grads}).
\revision{The result of which are} crisp decision boundaries and axis-alignment \rev{of the latent space. This results in an interpretable and disentangled representation.}
Secondly, we analyse the receptive field of the VAE's encoder in order to understand \rev{how the robot's gait phase is inferred}, see Sec.~\ref{section:encoders_receptive_field}.
\revision{Inference of the} gait phase is crucial for successful closed-loop planning in latent-space.
Thirdly, we compare the dynamic feasibility of the trajectories from the VAE-planner to those seen during training, see Sec.~\ref{section:evaluating_dynamic_feasibility} and show that \revision{the} VAE-planner's locomotion trajectories remain dynamically feasible \rev{as they are significantly varied}.
Next, we compare the distribution of realisable gait parameters output from \revision{the} VAE-planner to those seen in the dataset, see Sec.~\ref{section:generalisation_gait}.
We show the range of gaits achievable using our approach and compare this to the distribution seen during training.
Lastly, after showing that the approach can transfer from simulation to the real ANYmal B in \emph{Next Steps}, we push this limit and deploy the VAE-planner on ANYmal C without retraining, (Sec.~\ref{section:anymal_c}).
This experiment demonstrates the extent to which the VAE-planner is able to generalise to out-of-distribution scenarios.
In particular, ANYmal C exhibits significantly different dynamics to ANYmal B, which are detailed in Sec.~\ref{section:exp_design_deployment_anymal_c}.

\section{Related Work}

Planning and control for quadruped locomotion have advanced in leaps and bounds in recent years.
\revision{For example, hierarchical-optimisation frameworks split locomotion tasks into a series of smaller problems. Examples in this area include \emph{Dynamic Gaits} (DG)~\cite{bellicoso2018dynamic}, \cite{rathod2021Model} and \cite{grandia2021Multi}.}
DG and \cite{grandia2021Multi} enable a quadrupedal robot (such as ANYmal) to execute a wide variety of dynamic gaits (e.g.~trot, pace, lateral walk, jump) with real-time motion planning and control. However, to achieve this impressive range of behaviours, \revision{all these methods provide} each gait type with its own contact schedule and utilise an environment-specific footstep planner, ultimately limiting their capabilities.
\revision{The work in \cite{bledth2020Extracting} addresses the shortcomings of hierarchical planning approaches by learning a set of heuristic operating ranges in order to increase the overall dynamic range of quadruped locomotion. This is similar in philosophy to the work we present here. However, we achieve a broad dynamic range by learning a distribution over feasible trot gaits. This distribution is then sampled via our drive signals resulting in flexible and dynamic trot locomotion.  
This allows us to continually adjust the foot-swing timing of the airborne feet and is achievable much faster and to a broader degree than prior art. 
We distinguish this ability from methods which perturb the swing-duration of feet which are yet to break contact~\cite{rathod2021Model,Farshidian2017An,Mastalli2017Trajectory}. In addition, \cite{rathod2021Model} requires a heuristic metric for synchronising the contact scheduler with the current contact-state. Our approach sidesteps this requirement by performing closed-loop feedback directly in the learnt latent-space.}

Latent space approaches for planning and control learn useful and typically low-dimensional representations that can be used to control complex dynamics, without relying on known system models. Classic examples include \emph{Deep Variational Bayes Filters} (DVBF) \cite{DVBF} and \emph{Embed to Control} (E2C) \cite{embed2control}. DVBF produces dynamically consistent trajectories by traversing continuous paths in latent space whilst E2C learns a linear system model in which control problems can be solved. %
\emph{Conditional Neural Movement Primitives} (CNMP) \cite{CNMP} is a more recent latent space approach for robotic arms that generalises between a variety of tasks, such as pick-and-place and obstacle avoidance. 
Other recent works like UPN and PlaNet \cite{UPN,PlaNet} show impressive capabilities in simulation but are yet to be applied to real-world systems, including floating-base robots.

In the locomotion domain, \revision{Li et al. propose an approach in \cite{li2020Planning}, which utilise a learnt latent-action model to create different locomotion trajectories. However, this latent-action space does not capture robot dynamics, and samples trajectories via a random shooting method.}
\revision{In contrast}, \emph{First Steps} \cite{first-steps} learns a structured latent space based on feasible robot \emph{configurations} \revision{to capture the complete robot dynamics}. First Steps defines a set of performance predictors that can be used in an optimisation framework to control the robot. In practice, these performance predictors can be viewed as symbolic inputs (e.g.~`left front leg up') but drive the robot in continuous space. However, because \emph{First Steps} is trained on static snapshots of robot configurations, it does not learn from observable dynamics and thus requires more explicit structuring of the latent space
than is necessary. Our previously published work \emph{Next Steps}~\cite{NextSteps} addresses this shortcoming and significantly extends this framework to effective and robust closed-loop planning and control.
In this paper, we provide significantly more in-depth study into continuous variation of the gait parameters via planning in a structured latent-space.
In doing so, we further justify the utilisation of an oscillatory drive signal, analyse the quality of the VAE's trajectories, and push the limits of domain transfer through deployment on new robotic platform without retraining.

The \emph{Motion VAE} (MVAE) \cite{motionVAE} learns to represent dynamic trajectories in a structured latent space for the locomotion of computer-animated humanoids. This is similar to our emphasis here on learning representations for dynamic trajectories in the context of locomotion. However, moving from simulated to real physical systems, as is required for robotic applications, necessitates tackling additional complexities like latency, hard real-time requirements, and actuator dynamics.
In this work, we tackle these challenges and demonstrate that a single gait style contains sufficient richness to learn a structured latent space that can be exploited to manipulate gait characteristics that generalise beyond the range seen during training.
Unlike MVAE, our approach does not train on multiple gait styles, despite succeeding in producing them.

\revision{Other approaches build a model over multiple gait types and styles without utilising a structured latent-space. An example of which is~\cite{agrawal2021vision}. This utilises a vision-based system to sample a footstep schedule and gait for the terrain ahead. These form the input to an MPC approach. 
Alternatively, we choose not to condition the gait parameters on vision, but allow the operator to directly choose the gait style. Additionally, our system is able to generate a broad range of trot trajectories while being trained on a single demonstration of a trot gait with fixed parameters. Hence, we propose our method as an alternative to~\cite{agrawal2021vision} for creating a queryable model over different types of locomotion styles.}

Finally, a study conducted concurrently to our own \cite{yang2021fast} yields variation between gait types (walk and trot).
It utilises a reinforcement learning (RL) approach which employs a phase iterator similar to our drive signal.
However, this phase iterator is enforced, whilst our gait dynamics are discovered automatically purely from exposure to trot trajectories with constant parameters.

\section{Approach} \label{section:method}
\begin{figure}[t]
    \centering
    \includegraphics[width=1.0\linewidth]{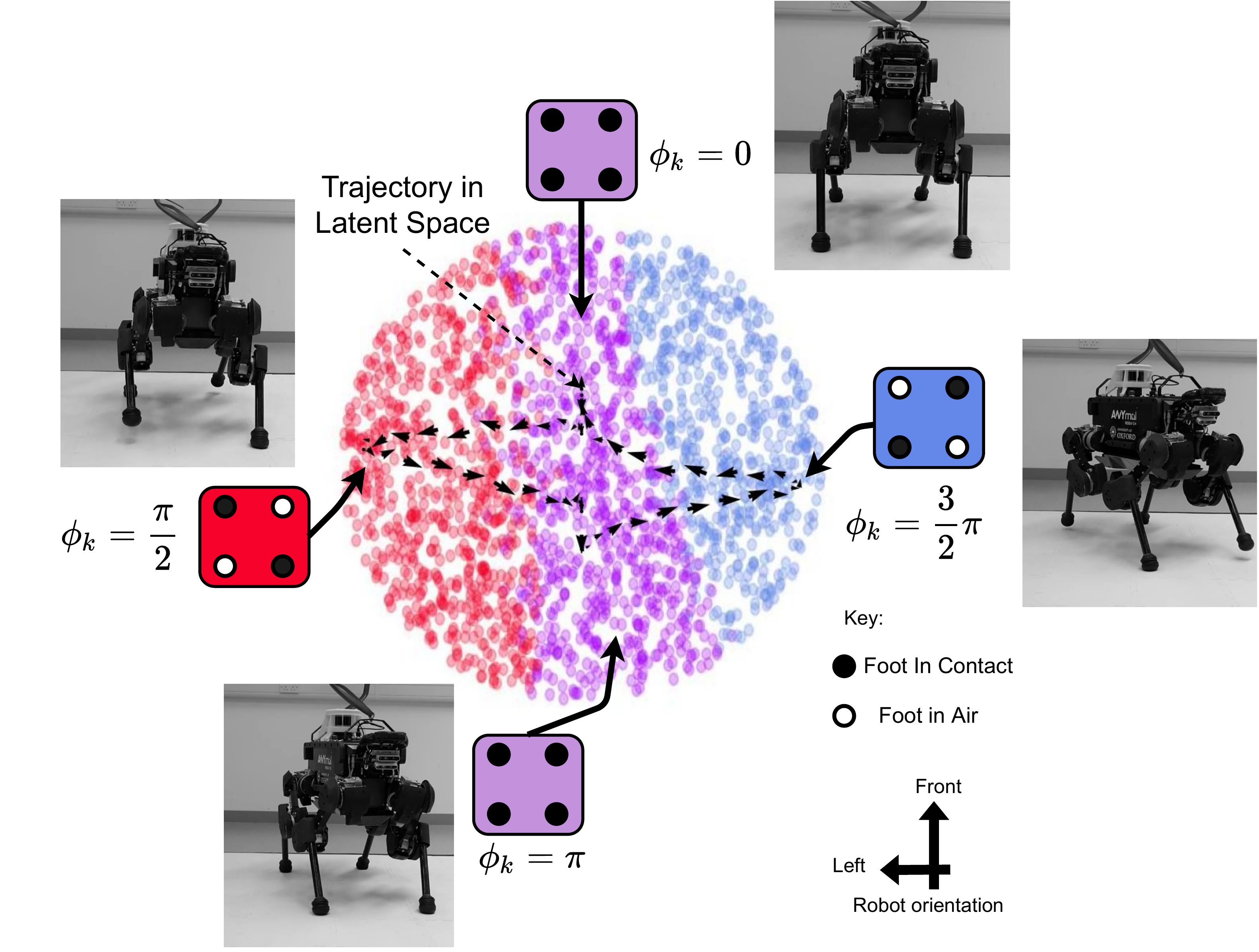}
    \caption{
    A slice through the structured latent space, colour coded to illustrate the ordering and clustering of the distinct stances which make up the trot gait.
    The component of the latent-space trajectory (black) along the horizontal axis contributes to the robot's footstep height, whilst the vertical component gives rise to the footstep length.
    Snapshots of the robot controlled using the VAE-planner illustrates the inter-play between these two latent dimensions.
    }
    \label{fig:latent-space}
    \vspace{-0.25cm}
\end{figure}

Our aim is to use unsupervised learning to infer a structured latent-space which facilitates real-time and smooth variation of key gait parameters. 
We conjecture that structure %
can emerge from the exposure of a suitable generative model to a gait with predetermined and constant characteristics such as cadence, swing height, and full-support duration.
Specifically, we propose that due to this structure, continuous latent trajectories result in robot locomotion, see Fig.~\ref{fig:latent-space}.
By inspecting this structure, we discover a disentangled latent-space where gait parameters are axis-aligned within this space (Sec.~\ref{section:exp_latent_space_structure}).
Periodic trajectories in latent space can then be decoded back to smooth robot locomotion as depicted in Fig.~\ref{fig:first_figure}.
Subsequently, the VAE is deployed as a planner in a real-time control loop. 

\textbf{VAE Architecture:} We train a VAE~\cite{vae,vae_1} to create a structured latent-space using observed dynamic data.
The input to the VAE $\rmX_k$ at time step $k$ consists of $N$ robot states sampled from simulated trot gaits with constant parameters (e.g. cadence and foot-step height).
These state-space quantities are values we wish either to control or are required to infer the gait phase.
These are joint angles; end-effector positions in the base frame; joint torques; contact forces; the base velocity; and the base pose evolution relative to a control frame, which is updated periodically.
These quantities are denoted as ${\rvx_k = [\rvq_k, \mathbf{ee}_k, \mathbf{\tau}_k, \mathbf{\lambda}_k, \mathbf{\dot{c}}_k, \Delta \rvc_k]}$, where $k$ is the time step.
Note that velocities and accelerations do not form part of the VAE's input, but are inferred from the input history.
This has dual benefits: first, it yields a lower-dimensional input space; and, second, it prevents sensitivity to fast-changing quantities such as the recorded joint accelerations during inference.

To deploy the VAE-planner in a closed-loop framework, we encode the input history at the control frequency $f_c$.
However, due to restrictions on the VAE's size caused by the tight computation bounds required for real-time control, the encoder input $\rmX_k$ is constructed using states spaced at a frequency of $f_{\enc}$:
\begin{equation}
    \rmX_k = [\rvx^\T_{k-r(N-1)}, \hdots, \rvx^\T_{k-r}, \rvx^\T_k], \label{eq:encoder_input}
\end{equation}
where $r=f_{c}/f_{\enc}$ is the ratio between the control and encoder frequencies. 
The input $\rmX_k$ is created every time step by sampling from an input buffer which stores every robot state $\rvx$ from time step $k$ to $k-r(N-1)$ at the control frequency.
We provide full details of the setting used in our experiments in Sec.~\ref{section:architecture_details}.

The VAE's decoder output $\hat{\rmX}^{+}_k$ predicts the current robot state $\rvx_k$ as well as $M$ future ones sampled at a frequency of $f_{\dec}=f_c$:
\begin{equation}
    \hat{\rmX}^{+}_k = [\hat{\rvx}^\T_k, \hat{\rvx}^\T_{k+1}, \hdots, \hat{\rvx}^\T_{k+M}] \revision{.}
\end{equation}
\revision{Here, $^{+}$ denotes future steps and $\hat{}$ denotes predictions.}

As the \emph{desired-feet-in-contact} is an input to the tracking controller, we also want to predict which of the four feet are in contact, $\rvs_k$, at the current time step, $k$, as well as $J$ steps in the future.
Inspired by \emph{First Steps}~\cite{first-steps}, we therefore utilise a feet-in-contact performance predictor $g_{pp}(\rvz_k)$. This is attached to the latent space, which estimates the probability of each foot being in contact:
\begin{align}
    \hat{\rmS}_k &= [\hat{\rvs}^\T_k, ..., \hat{\rvs}^\T_{k+J-1}]^\T 
\end{align}

To command the base twist of the robot, a high-level action command $\rva_k$ is utilised.
This represents longitudinal ($x$), lateral ($y$), and yaw ($\theta$) twist in the robot's base frame.
The latent state $\rvz_k$ and the action $\rva_k$ form the input to the decoder.

\textbf{Training the VAE:} 
We train the VAE and performance predictor together. The VAE's training loss is the modified ELBO formulation found in~\cite{betaVAE}.
This loss consists of a reconstruction loss (mean-squared error) plus the Kullback–Leibler (KL) divergence $\KL$ between the inferred posterior $q(\rvz|\rmX_k)$ and the prior $p(\rvz)$, multiplied by a hyper-parameter $\beta$:
\begin{align}\label{eq:loss_vae}
    \mathcal{L}_{\ELBO} = \MSE(\rmX^{+}_k, \hat{\rmX}^{+}_k) + \beta \KL[q(\rvz|\rmX_k) || p(\rvz)].
\end{align}

These ELBO terms are then summed with the binary cross-entropy loss between the predicted feet in contact and the recorded ones.
The latter term is scaled by $\gamma$, resulting in the overall loss
\begin{align}\label{eq:loss_full}
    \mathcal{L} = \mathcal{L}_{\ELBO} + \gamma \BCE(\rmS_k, \hat{\rmS}_k)
\end{align}\label{eq:losses_1}
The VAE training loss (Eq.~\ref{eq:loss_vae}), as seen in prior work \cite{betaVAE}, is responsible for any subsequent disentanglement found in the latent space.
The reconstruction error is weighed against the decomposition of the latent space using the hyper-parameter $\beta$.
This constraint encourages an efficient latent representation, containing only the required information for reconstruction, hence acting to regularise the latent space. 
As shown in \cite{betaVAE}, the $\KL$ term used with an isotropic unit Gaussian ($p(\rvz)=\mathcal{N}(\mathbf{0},\rmI)$) encourages conditional independence within $\rvz$.

In our approach, as well as that of \emph{First Steps}~\cite{first-steps}, a structured latent space is encouraged by backpropagating gradients from the performance predictor's loss through to the encoder input.
However, we hypothesise that useful structuring of this space is inferred from the continuous trajectories used for training, and, in contrast to \emph{First Steps}, no explicit labelling for each stance is required.

\begin{figure}[t]
    \centering
    \includegraphics[width=1.0\linewidth]{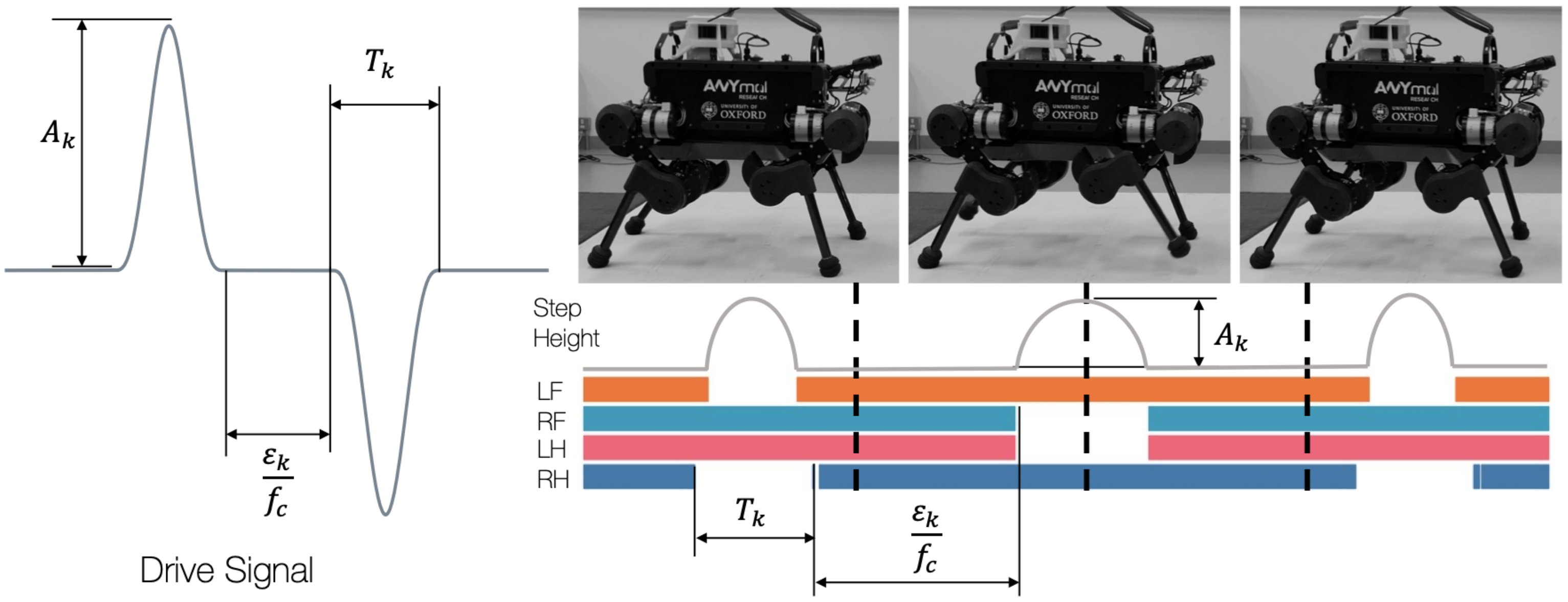}
    \caption{
    An oscillatory \emph{drive signal} overwrites the dimension with the smallest variance in the structured latent-space.
    The amplitude $A_k$, time-period $T_k$ and stance duration counter $\epsilon_k $ of this signal control the robot's foot swing height, cadence and full-support duration in real time.
    }
    \label{fig:annotated_drive_signal}
    \vspace{-0.25cm}
\end{figure}

\subsection{Control over the gait parameters}\label{section:drive-signal}

Once the VAE has been trained as described above, we find that the learnt latent-space is disentangled.
We discover that an oscillation injected into one latent dimension decodes to the robot taking steps.
Further analysis detailed in Sec.~\ref{section:exp_latent_space_structure} finds that adding a second oscillation into the latent space varies the robot's step length.
The former oscillation we denote as the \emph{drive signal}, and the latter as the \emph{trot signal}.
We find that in order to control the robot, we only need to inject the drive signal into latent space, and can infer the trot signal.
By modulating and visualising the joint-space output from the decoder, we discover that varying the amplitude and phase of the drive signal leads to continuously variable trot locomotion.
Therefore, we choose a specific drive signal with features which map directly to the gait parameters we wish to control.
These parameters are the robot's cadence, stance duration and step height.

While any periodic oscillation (e.g. $sin$) decodes to robot locomotion, we choose a drive signal featuring specific components allowing for control over the robot's cadence, swing duration and step height.
The drive signal chosen here is a modified $\sin^3$ oscillation with amplitude $A_k$, and phase $\phi_k$:
\begin{align}\label{eq:drive_signal}
  \rvz_{k,d_z} =  A_k \sin^3(\phi_k).
\end{align}
The amplitude $A_k$ controls the foot swing height, and the phase $\phi_k$ governs cadence and support duration.

To control the robot's swing and stance duration separately, we set the drive signal's time-period $T_k$ and we employ a stance counter $\epsilon_k$.
The time-period $T_k$ is equal to the swing duration, whilst the time that the drive signal is equal to zero is extended by $\epsilon_k$ time steps to introduce a full-support duration of \SI[parse-numbers = false, number-math-rm = \ensuremath]{(\epsilon_k / f_c)}{\second}.
Hence, once both $T_k$ and $\epsilon_k$ are used together, the phase dynamics are:
\begin{align}\label{eq:phase_dynamics}
    \phi_{k+1} = 
    \begin{cases}
      \phi_k & \text{if $\phi_k \bmod \pi = 0$ and $k_{\epsilon} < \epsilon_k$} \\
      \phi_k + {2\pi}/{T_k} & \text{otherwise}
    \end{cases}
\end{align}
and, in tandem, the counter $\epsilon_k$ is updated as:
\begin{align}\label{eq:stance_dynamics}
    k_{\epsilon} \gets 
    \begin{cases}
      k_{\epsilon} + 1 & \text{if $\phi_k \bmod \pi = 0$ and $k_{\epsilon} < \epsilon_k$} \\
      0 & \text{otherwise}
    \end{cases}
\end{align}

\revision{A $\sin^3(\cdot)$ drive signal is chosen as it is smooth over its domain meaning that the decoded trajectories will also be smooth. Additionally, a $\sin^3(\cdot)$ drive signal's gradient is zero at $\phi_k=K\pi$ for $K \in \mathbb{Z}$. 
Therefore, there is a continuous transition between the $\sin^3(\cdot)$ part of the drive signal and the points where the drive signal is held artificially zero for $\epsilon_k$ control-ticks.
The resulting signal is shown in Fig.~\ref{fig:annotated_drive_signal} along with the decoded contact schedule and swing trajectory.
}

\subsection{Planning for closed-loop control}\label{section:closed-loop}

Once the VAE is trained, it is fast enough to act as a planner in a closed-loop controller.
Thus, our approach can react to external disturbances and mitigate against real-world effects such as unmodelled dynamics and hardware latency.
For closed-loop control, we begin by encoding a history of robot states from the raw sensor measurements to infer the current gait phase.
We store a buffer of past robot states and sample from this at $f_{\enc}$ to create the encoder's input.

With an estimate of the current latent variable, we overwrite latent dimension $d_z$ with the drive signal (see Sec.~\ref{section:drive-signal}).
Next, we employ a second-order Butterworth filter to smooth the latent trajectory and further smooth the locomotion plan.
In essence, the drive signal encourages the decoder to output the next open-loop prediction while the other latent variables infer the gait phase from the raw sensor input.
This process is repeated at the control frequency (\SI{400}{\hertz}).

The latent variable $\rvz_k$ and a desired base twist $\rva_k$ are decoded to produce ${\hat{\rmX}^+_k=g_{\dec}(\rvz_k, \rva_k)}$.
From this, the joint-space trajectory $\hat{\rmQ}_k$, and local base velocity $\hat{\dot{\rmC}}_k$ are extracted and derived or integrated to produce the base and joint positions, velocities and accelerations.
These quantities and the predicted contact schedule are sent to the whole-body controller (WBC)~\cite{WBC}.
The WBC solves a hierarchical optimisation problem to calculate the joint torques which are commands sent to the actuators.
The series of constraints enforced by the WBC are: contact creation, friction constraints and torque limits.
Next, the WBC applies forward kinematics to the VAE's trajectory to track it in task-space.
Note that the WBC does not compensate for infeasible plans, i.e. the VAE's trajectories shown in Sec.~\ref{section:sliders} are dynamically consistent otherwise the robot fails to walk.

\subsection{Disturbance detection and response}\label{section:methods_disturbance_handling}
Our approach is able to both detect and react to disturbances. The VAE is trained using canonical feasible trajectories. Therefore, any disturbances are characterised as out of distribution with respect to the training set.
Given the generative nature of our approach, this discrepancy is quantified during operation by the trained model via the Evidence Lower-Bound (ELBO, Eq.~\ref{eq:loss_vae} where $\beta$ is set to one). We will show in the evaluation (Sec.~\ref{section:disturbance_rejetion}) that even a rudimentary response strategy serves to increase the range of disturbance the system can reject.

\section{Implementation Details}

In order to deploy the VAE-planner on-board the ANYmal quadruped, there are a number of real-world constraints which affect the VAE-planner.
The first of which is that in order to deploy the VAE in a real-time control loop, there is a constraint on the VAE's inference time.
Secondly, we discuss how we address the simulation to reality gap.
We discuss the VAE's specific architecture, the hyper-parameters used to train the model, and finally, specific the specific criteria required for domain transfer.

\subsection{Dataset generation}\label{section:dataset_generation}
To train the VAE and create the structured latent space, we require a set of continuous trot trajectories.
\revision{We restrict these trajectories to quadruped locomotion over flat ground \emph{only} so that we can study and understand how to generate versatile locomotion utilising a structured latent-space.}
\revision{The training trajectories are generated using \emph{Dynamic Gaits} (DG)~\cite{bellicoso2018dynamic}.}
DG is a hierarchical planning and control framework which is used with a \emph{fixed} contact schedule and predefined footstep heights.
The swing, full-stance durations and footstep heights set to \SI{0.5}{\second}, \SI{75}{\milli\second} and \SI{0.10}{\meter}, respectively.

DG is made up of a footstep planner, a base motion planner and a whole-body controller (WBC).
The footstep planner computes the next four steps over the gait period using an inverted-pendulum model.
The base motion planner \revision{solves for} the base trajectory over the gait period using a centroidal dynamics model~\cite{centroidal_dynamics}.
The latter is constrained with a Zero-Moment Point (ZMP)~\cite{ZMP} criterion, \revision{the footstep positions and schedule from the footstep planner.} 
The WBC~\cite{WBC} outlined in Sec.~\ref{section:closed-loop} converts the task space trajectories to joint feedforward torques, reference positions and velocities: these are sent to the actuators.

The dataset is generated by uniformly sampling desired base twist and executing DG in the \emph{RaiSim} physics simulator~\cite{raisim}.
\revision{The fidelity of the simulation is improved by modelling the dynamics of the} Series-Elastic Actuators (SEA)~\cite{SEA} in the ANYmal's joints using an \emph{actuator network}~\cite{hwangbo2019learning}.
\revision{We specifically utilise the network found in \cite{gangapurwala2020rloc}}.  This network takes into account the commanded positions, velocities, feed-forward torques, and low-level PD gains
The actuator network is essential for good performance as the input response of SEAs depends on a history of states, inputs, and the low-level control law.

\subsection{VAE architecture details}\label{section:architecture_details}

Fig.~\ref{fig:first_figure} outlines the approach's architecture, and here we describe the VAE's specific details. 
Prior to the ablation study in Sec.~\ref{section:ablation_study}, the VAE's encoder, decoder and stance performance predictor have two hidden layers and widths of 256 units, using ELU non-linearities~\cite{elu}. 
The encoder input is created using $N=80$ robot states sampled at \SI{200}{\hertz} -- representing a history of \SI{0.4}{\second} -- from the encoder input, which is of size $5120$ units.
The input is compressed via a latent space of $125$ units which is concatenated with an action of $3$ units.
Next, the decoder outputs the current state and the next $M=19$ robot states at the control frequency of \SI{400}{\hertz} (preview horizon \SI{47.5}{\milli\second}, output size: $1216$ units). 
The performance predictor predicts the current feet in contact and two future states.
Finally, hyper-parameters used for training are $\beta=1.0$, $\gamma=0.5$, with a learning rate of \num{1e-3} using the Adam optimiser.
Training is terminated after \num{1e+6} gradient steps.

\subsection{Domain transfer}\label{section:domain_transfer} 
To achieve domain transfer and deployment on the real robot, some modifications are required.
The contact forces are artificially set to zero when the robot measures no contact (as determined by a probabilistic contact estimator)~\cite{ANYmal}.
This is necessary since the real-world ANYmal robot measures large contact forces even during swing motion, as these forces are inferred from torque residuals and are affected by model error. In contrast, simulators estimate no contact force during swing.
As mentioned in Sec.~\ref{section:closed-loop}, a Butterworth filter with a cutoff frequency of \SI{10}{\hertz} is employed to smooth the latent trajectory.
Due to the strict computation budget for real-time control, inference times for the VAE are restricted to at most \SI{1}{\milli\second}, which imposes constraints on the capacity of the model (see Sec.~\ref{section:ablation_study} for an ablation of model hyperparameters).
Our largest model takes approximately \SI{1}{\milli\second} for the VAE computation, which is roughly equal to the computation time of the WBC.
\section{Experimental Design}

In this section, we explain how the latent-space properties are discovered and how we assess the performance of the VAE-planner deployed on the real robot.
In doing so, we motivate the following guiding questions.
The aim of these questions is to analyse the latent space once the VAE is trained and to analyse the capabilities of the VAE as a flexible and robust locomotion-planner.

We investigate
(i) the structure induced in the latent space (Sec.~\ref{section:exp_latent_space_structure}), 
(ii) the effect backpropagating the binary cross-entropy gradients through the encoder has on the latent-space structure (Sec.~\ref{section:bce_grads})
(iii) the sensitivity of our approach to variations in key hyper parameters (Sec.~\ref{section:ablation_study}), 
(iv) which parts of the encoder's input are used to infer the robot's gait phase (Sec.~\ref{section:encoders_receptive_field}),
(v) to what extent the locomotion parameters can be varied online (Sec.~\ref{section:sliders}), 
(vi) the feasibility of the locomotion plans produced (Sec.~\ref{section:evaluating_dynamic_feasibility}), 
(vii) a comparison between the gait parameters seen during training and those shown in experiments using the VAE-planner (Sec.~\ref{section:generalisation_gait}), 
(viii) the degree to which disturbance detection, coupled with a rudimentary recovery strategy, further increases the robustness of our approach (Sec.~\ref{section:disturbance_rejetion}), and finally, 
(ix) whether the VAE-planner can be deployed successfully on the next generation ANYmal C robot without retraining (Sec.~\ref{section:anymal_c}). 
Please see the following video for an extended set of experiments along with a brief description of our approach (\url{https://youtu.be/GT2WLh2Ackc}).

\subsection{Investigating the latent space}\label{section:methods_structure_latent_space}
We wish to examine if there is any structuring in the latent space as well as investigate if any locomotion properties are disentangled within.
This knowledge is crucial in understanding how to solve for locomotion trajectories in latent space.

\textbf{Latent Space Structure:}
Structure in latent space manifests itself as clustered latent variables.
During training, we expect that points in latent space which are of the same gait type will become gathered together.
To verify this, we sample a set of random latent variables which we pass through the stance performance predictor.
These points are plotted and colour-coded based upon their predicted stance.
The resulting plot can be found in Sec.~\ref{section:exp_latent_space_structure}, specifically in Fig.~\ref{fig:normal_latent_space}.

\textbf{Latent Space Disentanglement:} 
We wish to see what trajectories in latent space look like, and if any of the dimensions within are interpretable.
State-space trajectories of the robot trotting from the test set are encoded into latent-space, and the subsequent latent-space paths are visualised.
These paths in latent space are oscillatory and each oscillation has its own phase.

In order to understand what each oscillation encodes, we artificially inject sine waves into each latent dimension in turn.
Decoding these trajectories and visualising the joint-space paths reveals that the latent space is indeed disentangled.
See Sec.~\ref{section:exp_latent_space_structure} for full details.
This revelation informs how we solve for locomotion paths in latent space.

\subsection{Backpropagating the binary cross-entropy through the encoder}
As in \emph{First Steps}~\cite{first-steps}, we encourage the latent space to become structured by backpropagating the BCE loss for the contact state of the feet through the encoder.
We train an alternative VAE where we detach the BCE gradients at the latent space.
This means that the latent-space structure is not informed by the prediction loss of the feet in contact performance predictor.
The resulting latent-space of the VAE with detached gradients is compared to the original.
Firstly, we look for the crispness of the decision boundaries between stance clusters. 
Secondly, we look to see if the decision boundaries are aligned with any dimensions in latent space. This analysis is aimed to investigate the importance of using the contact performance predictor to generate feasible locomotion.

\subsection{Analysing the encoder's receptive field}
In our experiments on the real robot, the VAE is able to infer the robot's gait phase from raw sensor inputs.
We wish to understand which parts of the encoder's input are utilised to infer this.
We utilise activation maximisation (AM)~\cite{activation_maximisation} to measure the flow of gradients through the encoder.
This procedure requires backpropagating from some randomly sampled input encoding until the encoder output matches a predetermined target value.
The target is chosen to match key points along the robot's gait phase and are chosen such that the drive signal's phase is equal to $0.0, \pi/2, \pi/4, \pi/8, \pi$.
The gradients at the encoder input resulting from this optimisation are recorded.
The magnitude of these values reveal which parts of the input are required to infer the gait phase.

\subsection{Studying the VAE's sensitivity to key hyper-parameters}
The VAE's architecture is ablated and the VAE-planner is tested in simulation until the gait phase can no longer be inferred.
This leads to noisy and jerky trajectories in simulation.
Firstly, the size of the latent dimension is reduced from $125$ to a minimum of $6$ in strides of $32$ units.
Secondly, the number of input states $N$ is reduced from $80$ units to $60$, sampled at \SI{200}{\hertz}, meaning that the encoded input history reduces from \SI{0.4}{\second} to \SI{0.3}{\second}.
Thirdly, the encoder frequency is halved to \SI{100}{\hertz} with $N=80$.
Lastly, the widths of the encoder, decoder and performance predictor are reduced in increments of $32$ units from $256$ to a minimum of $64$.

\subsection{Varying the gait parameters online}
We wish to vary the robot's gait parameters continuously whilst the robot is walking. 
This is straight-forward for the robot to do.
We utilise a simple ROS publisher with which sets the drive signal's amplitude, time-period, and stance duration.
The robot's base twist is commanded independently using the action $\rva_k$. 
In fact, we utilise a navigation waypoint following controller which produces the action so that the robot walks around a square trajectory in our lab.

\subsection{Analysing the dynamic feasibility of the VAE's trajectories}

The previous guiding question asks how broad the set of manoeuvres from the VAE-planner are.
We now compare the dynamic feasibility of these trajectories to plans from DG.
Please note that this dynamic feasibility comparison is undertaken as a post-experiment analysis.
On the robot, the output of the VAE-planner is sent directly to the WBC unaltered.

To evaluate the dynamic feasibility of trajectories synthesised by our VAE, we measure the distance of the Zero Moment Point (ZMP)~\cite{ZMP} to the support line (i.e. when one pair of legs are swinging) and then compare this distance with that in the synthetic dataset.
The ZMP is a commonly used criterion in model-based legged robot control~\cite{ZMP,bellicoso2018dynamic}.
This analysis is performed for the trajectories in the dataset as a baseline to which we can compare.
Dynamically stable trajectories have the ZMP as close as possible to the support line.

\begin{figure*}[t]
    \centering
    \includegraphics[width=1.0\textwidth]{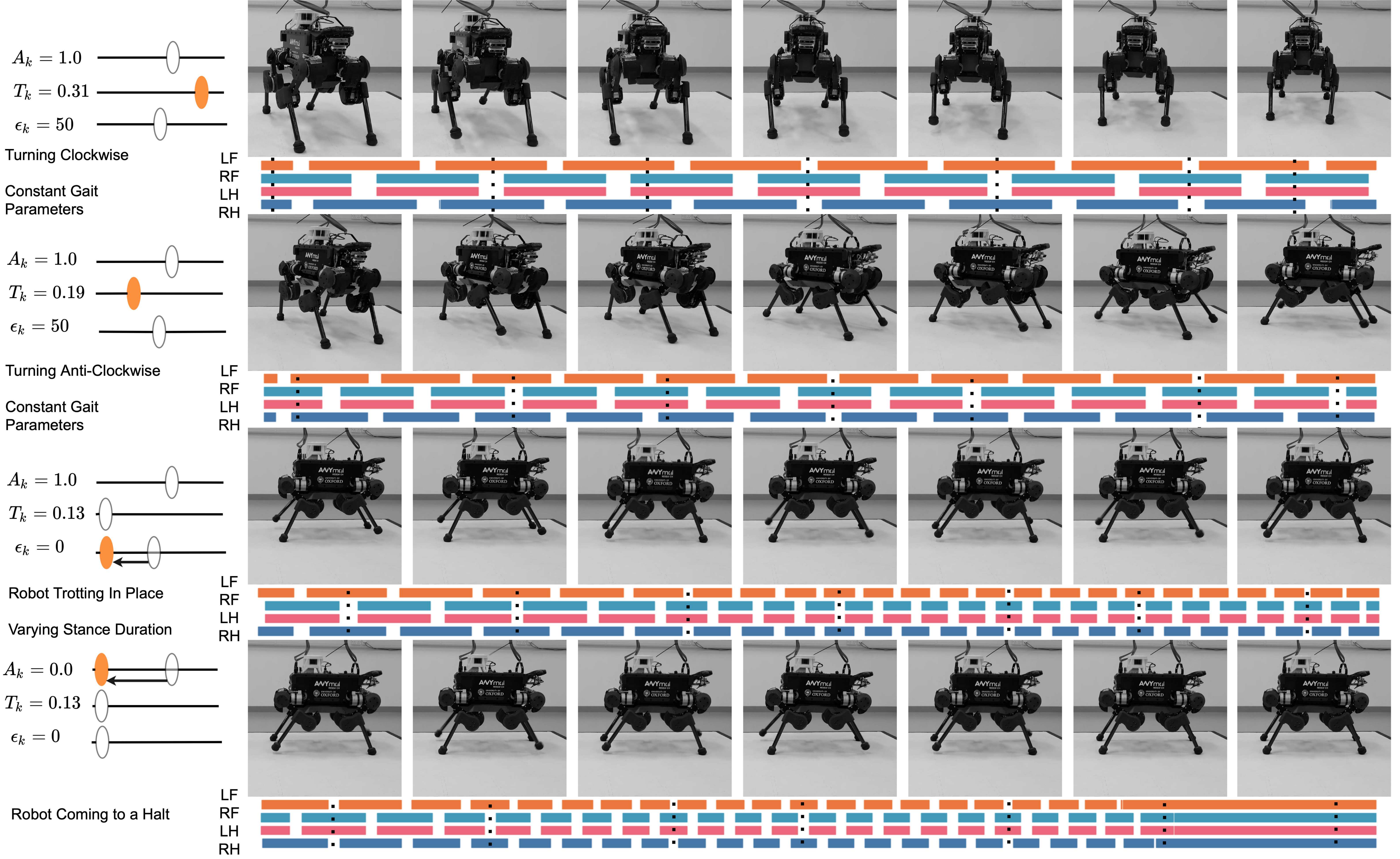}
    \caption{Closed-loop control of the real ANYmal quadruped using our VAE-planner. This demonstrates user-controlled variation of gait parameters on the fly.
    Here, coloured rectangles represent the full-stance phase, whilst white space denotes the swing duration.
    The top row shows a trot gait with an introduced quadrupedal stance phase (gait cycle of \SI{0.75}{\second} -- swing \SI{312.5}{\milli\second}, stance \SI{62.5}{\milli\second}; a gait cycle consists of a swing phase for each of the leg pairs).
    Next, the swing duration $T_k$ is reduced using the time-period slider (gait cycle of \SI{0.5}{\second} -- swing \SI{188}{\milli\second}, stance \SI{62.5}{\milli\second}).
    The third row illustrates the effect of reducing the stance duration counter $\epsilon_k$ to produce trot with reducing full-stance phases (gait cycle of \SI{250}{\milli\second} -- swing \SI{125}{\milli\second}, stance \SI{0.0}{\second}).
    Finally, transition into standing occurs when the drive signal amplitude $A_k$ is reduced to zero.
    To view the full range of movements, please see the following video \url{https://youtu.be/GT2WLh2Ackc}.
    }
    \label{fig:cadence-comparison}
    \vspace{-0.15cm}
\end{figure*}

\subsection{Measuring the distribution of gait parameters achievable with the VAE}

The gait parameters set during the robot experiments are recorded meaning that we can measure the distribution of visited states.
To show the entire range of movements and compare against those seen during training, we plot a box and whisker plot of the stance and swing durations.
This shows the ability of the VAE-planner to generalise producing gait parameters not seen in the training distribution.

\subsection{Detecting and reacting to disturbances}

We wish to analyse the degree to which disturbances can be detected and whether increasing the robot's cadence helps the robot to recover from them.
As mentioned, we monitor the ELBO as this is a measure of the evidence for an encoded input relative to the learnt distribution.
Hence, a large spike in this value (given the formulation in Eq.~\ref{eq:loss_vae}) is a consequence of a disturbance.

To perturb the robot, we utilise a push broom and disturb the robot's base.
If the ELBO value surpasses a pre-determined constant, we characterise the event as a disturbance.
The ELBO value is calibrated such during normal operation, the ELBO remains below this "disturbance" threshold.
This is easily found by walking the robot using the VAE-planner for \SI{2}{\min}, whilst recording the ELBO.

Following disturbance detection, the VAE-planner automatically reduces the drive signal's time-period to increase the robot's cadence.
This response is inspired by human locomotion in response to slippage~\cite{Moyer2006GaitPA}: a group of participants encounter a slippery surface and their reaction is recorded, resulting in an increase in cadence.

We evaluate the effectiveness of increasing the robot's cadence by repeating the push experiment with and without the cadence increase.
The base velocity is utilised to measure the size of the disturbance.
Larger velocities arise from bigger pushes.
We make a comparison between the reactive VAE-planner and a constant cadence version.

In addition, the latent-space trajectory is inspected during and after a disturbance.
We are interested in how this is affected by out of distribution occurrences like a shove from a push broom or a kick to the robot's base.

\subsection{Deployment on ANYmal C without retraining}\label{section:exp_design_deployment_anymal_c}

The VAE which is trained using simulated ANYmal B locomotion data is deployed on ANYmal C without retraining.
Since our generative model has successfully transferred from simulation to the real robot, a key question is can the VAE-planner work on a different robot with similar kinematics, but significantly different dynamics?
Therefore, we deploy the VAE-planner on a different robot which has a few key differences.
Firstly, the ANYmal C has around twice the torque limit of ANYmal B, \SI{80}{\newton \meter} and \SI{40}{\newton \meter} respectively.
Secondly, ANYmal C is significantly heavier than ANYmal B weighing in at \SI{50}{\kilo \gram} to B's \SI{35}{\kilo \gram}. 
Finally, ANYmal C's actuators have a lower bandwidth due to the additional reduction gearing. 
The only alterations made to the VAE-planner are firstly the torques are standardised to reflect the difference in peak torque demand, and secondly, the WBC's internal dynamics' model is updated to ANYmal C.

As previously discussed, the VAE-planner can be utilised to measure differences between the learnt distribution and the encoded one.
Therefore, this approach is able to characterise differences between the two robots.
Again, we utilise the ELBO for this, and we record this quantity when the VAE-planner is deployed on ANYmal C.
The ELBO distribution can be separated in the KL divergence and the reconstruction error.
The former provides insight into the distribution of encoded values and the latter acts as a prediction loss.

To verify if the differences in KL-divergence collected from both robots are statistically significant, we utilise a Mann-Whitney U-test~\cite{mann1947}.
Our null hypothesis is that the median of the distribution from ANYmal B is equal to the median from the distribution from ANYmal C.
The values of KL-divergence are collected from deploying the same VAE-planner on both robots.
In total, we analyse $2.5\times 10^4$ values from each robot, which is equivalent to \SI{62.5}{\second} at \SI{400}{\hertz}.
The null-hypothesis is rejected if the resulting p-value is less than our chosen statistical significance value of \SI{0.1}{\percent}.
This procedure is repeated for the two sets of reconstruction error gathered from each robot.

\section{Experimental Results} \label{section:evaluation}

Now that we have posed our guiding questions and stated how to answer them, we present the results.
These questions can be split up into two categories: (a) introspection of the VAE and (b) analysis of the VAE as a planner deployed on two real-world platforms. 
Please see the following video for a complete set of experiments along with a brief description of our approach: \url{https://youtu.be/GT2WLh2Ackc}.

\subsection{Structure induced in the latent space}\label{section:exp_latent_space_structure}
The latent space is inspected to discover what structure exists and if any locomotion properties are disentangled within.

\begin{figure}[t]
    \centering
    \includegraphics[width=1.0\linewidth]{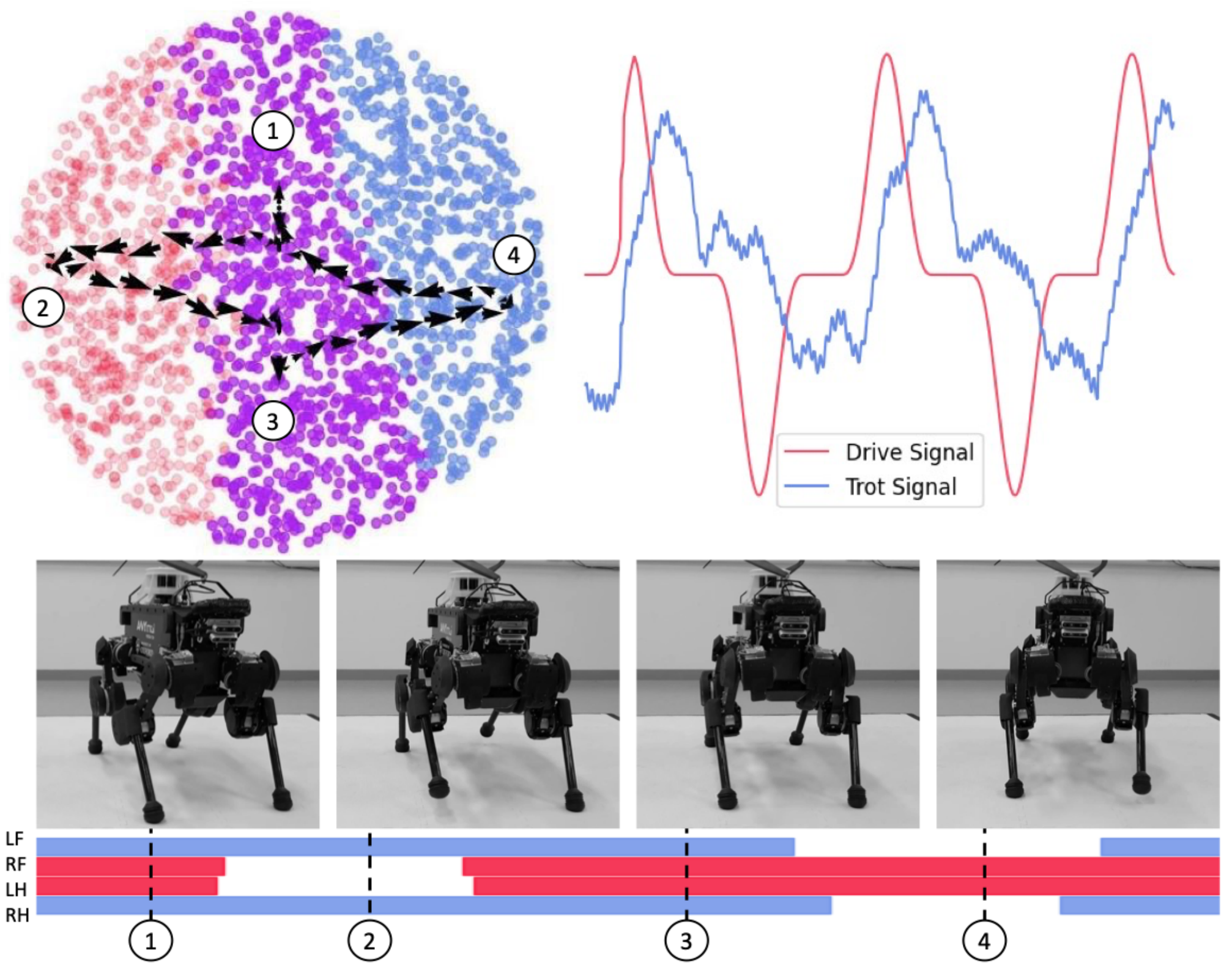}
    \caption{The latent space and trajectory in black during unperturbed operation. The latent space trajectory is result of the combination of the red drive-signal along the horizontal dimension and the blue trot-signal in the vertical. The drive signal is user controlled, whilst the trot-signal is inferred. The subsequent robot locomotion and contact schedule is included.}
    \label{fig:normal_latent_space}
\end{figure}

\textbf{Latent Space Structure:}
Fig.~\ref{fig:normal_latent_space} shows samples from the latent space colour-coded by their \emph{predicted} stance.
For trot, there are four stances: the full-support phase, left front and right hind in contact, another support phase and finally right front plus left hind in contact.
Fig.~\ref{fig:normal_latent_space} reveals that the latent space has emerged clustered by stance and that, due to the ordering of these stances, a periodic trajectory decodes to a trot gait.
This favourable structure is inferred from the continuous trot trajectory input during training.

\rev{The ordering of the clusters matches the sequence of the trot gait. The reason for this is that the encoder's input contains information about the evolution of the robot's state over time. This additional information coupled with the training paradigm means that continuous trajectories in latent space match with continuous trajectories in state space. This is in contrast to the work in \emph{First Steps}~\cite{first-steps}. \emph{First Steps} generates a crawl gait via gradient descent in a structured latent-space. However, the input to that model is a static snapshot of the robot's state with no temporal information (e.g. $\rvx_k$). Since there is no information about the evolution of the robot's state in the encoder's input, additional structuring was required to order the clusters in the crawl gait sequence.}

\textbf{Latent Space Disentanglement:}
By examining the latent variables, we discover that oscillations injected into just two dimensions in the latent space decode to continuously varying trot trajectories. This result stems from a latent space where variation in footstep length and cadence is aligned along one dimension, while variation in footstep distance lies along another.
Specifically, the time period of what we denote the \emph{drive signal} oscillation controls the robot's cadence, whilst its amplitude is proportionate to the footstep height.
In addition, the amplitude of the second signal, which is $\pi/2$ out of phase with the drive signal, controls foot swing length and as such is denoted as the \emph{trot signal}.
Given that other work has tried to explicitly build this structure into locomotion systems~\cite{yang2021fast}, it is important to emphasise that this disentanglement \emph{emerges} in our study as a result of the training paradigm and data. 
Only the synthetic drive-signal needs to be injected into the latent space for closed-loop control; the trot signal is inferred.

\textbf{Visualising The Latent-Space Trajectory:}
We plot the injected drive-signal and inferred trot signal in Fig.~\ref{fig:normal_latent_space} in red and blue respectively.
When these two signals are plotted against one another, they combine to form a cycle in latent space.
This is plotted as the black-vector field in Fig.~\ref{fig:normal_latent_space}.
This cycle is annotated to show how the latent-space trajectory visits each stance cluster in turn.
We also show the corresponding contact schedule with images of the robot in their matching configurations.

To begin at point (1), the robot has four feet in contact, before moving anti-clockwise tracing out a triangular lobe in the red region to point (2).
This lobe forms the first foot-swing. 
The trajectory continues anti-clockwise through the magenta region (3) before tracing another lobe through the blue area back to point (1).
This cycle repeats as the robot takes more steps.

\begin{figure}
    \centering
    \includegraphics[width=0.9\linewidth]{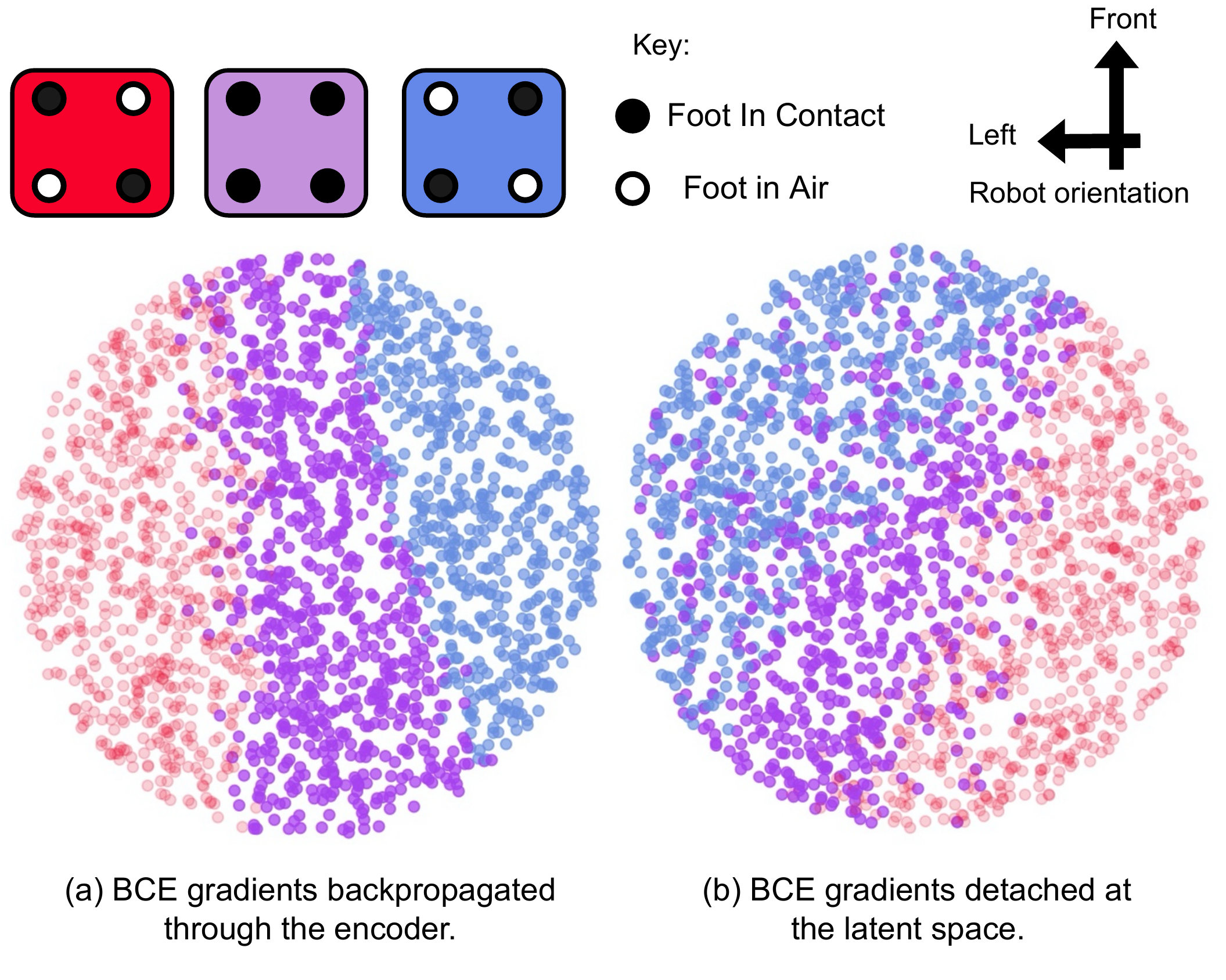}
    \caption{A comparison of a latent space actively structured using BCE gradients which flow through the encoder's input (sub-figure (a)) to the vanilla case, where BCE gradients are detached at the latent space (sub-figure (b)). Points in latent space are coloured by their stance. In the key above the latent-space images, a filled-in circle represents a closed contact and a circular outline denotes an open contact.}
    \label{fig:bce_gradients_ls}
\end{figure}

\subsection{Backpropagating the BCE loss through the encoder}\label{section:bce_grads}

We compare the latent space created by backpropagating the BCE gradients through the encoder to the vanilla case, where these gradients are detached before the latent space in Fig.~\ref{fig:bce_gradients_ls}.
The latent space structured using the BCE gradients exhibits crisp decision boundaries for the stance clustering.
These boundaries are aligned perpendicular to the drive-signal dimension in latent space, meaning that stance classification results mostly from this dimension.
In contrast, the detached gradient latent-space has fuzzy decision boundaries.
This is less than ideal as the predicted contact state is an input to the controller, and vitally affects the propagation of the dynamics model.
If this is a noisy prediction, the robot will change contact state in an unpredictable way.
Note that the decision boundaries in the detached case are rotated at roughly \ang{30} in comparison the nominal case. 
This results in the latent space being no longer axis aligned with the drive signal. 
In order to create locomotion trajectories and have independent control over the gait parameters, namely step height and length, the drive signal needs to be rotated to align with the decision boundaries.
This is not required for the nominal case, where the BCE gradients are backpropagated through the encoder. 
Therefore, the crisp decision boundaries and the axis-alignment justifies our decision -- and more crucially highlights the importance -- of using the BCE gradients to structure the latent space.

\subsection{Sensitivity To Hyper Parameters}\label{section:ablation_study}

As mentioned in Sec.~\ref{section:domain_transfer}, real-time performance is only achievable if VAE inference can be performed in under a \SI{1}{\milli \second}.
Therefore, we report the results of our ablation study where the VAE's channel capacity is systematically reduced.
Table~\ref{table:success_results} summarises the results of this study.
Firstly, the latent dimension is reduced from $125$ to a minimum of $6$.
The VAE with latent size of $6$ units is deployed successfully in simulation and on the real robot.
Next, the VAE's width is reduced in 32-unit increments and a limit of $128$ units is found.
This corresponds to a reduction in channel capacity by \SI{52.8}{\percent}.

Additionally, the window of time used to construct the VAE's input is reduced from \SI{0.4}{\second} to \SI{0.3}{\second} whilst maintaining an encoder frequency of \SI{200}{\hertz}.
This results in poor open-loop performance as the VAE is no longer able to learn the gait phase.
This is a result of the swing duration in the dataset which is \SI{0.45}{\second}.
Therefore, an input history over the last \SI{0.3}{\second} is insufficient to capture the gait phase.
Next, we investigated reducing the encoder's sampling frequency $f_{\enc}$ by halving it to \SI{100}{\hertz} whilst the history remains sampled over \SI{0.4}{\second}. 
Though this speeds up inference, the resulting trajectories are less smooth than the \SI{200}{\hertz} encoder and the robot is not stable during closed-loop operation.

As a result of these analyses, the minimum VAE architecture requires a latent space of $6$ units, an input history of \SI{0.4}{\second} sampled at \SI{200}{\hertz}, and a hidden layer size of $128$ neurons.
We confirm this by deploying this reduced model on the real robot.

\begin{table}[h]
\vspace{-0.1cm}
\centering
\caption{We perform an ablation study to find the minimum channel capacity required for the VAE-planner to transfer successfully to the real robot. 
The original VAE model denoted as (O) is found in the bottom right-hand corner of this table.
The latent-space size is reduced to six whilst the width remains 256. The width is reduced to 64 in units of 32. The smallest model which transfers to the real robot has width of 128 and latent size of six.}
\label{table:success_results}
\begin{tabular}{c | c c c c c}
\multicolumn{1}{c}{}  & \multicolumn{5}{c}{Model Width} \\ %
\multicolumn{1}{c}{ Latent Size} & \multicolumn{1}{c}{64} & \multicolumn{1}{c}{96} & \multicolumn{1}{c}{128} & \multicolumn{1}{c}{192} & \multicolumn{1}{c}{256}
\\ \hline 
 6  & \textbf{F} &  \textbf{F} &  \textbf{T} & \textbf{T} & \textbf{T} \\
29  & \textbf{-} &  \textbf{-} &  \textbf{-} & \textbf{-} & \textbf{T} \\
61  & \textbf{-} &  \textbf{-} &  \textbf{-} & \textbf{-} & \textbf{T} \\
125 & \textbf{-} &  \textbf{-} &  \textbf{-} & \textbf{-} & \textbf{O} \\
\end{tabular}
\end{table}

\begin{figure}
    \centering
    \includegraphics[width=1.0\linewidth]{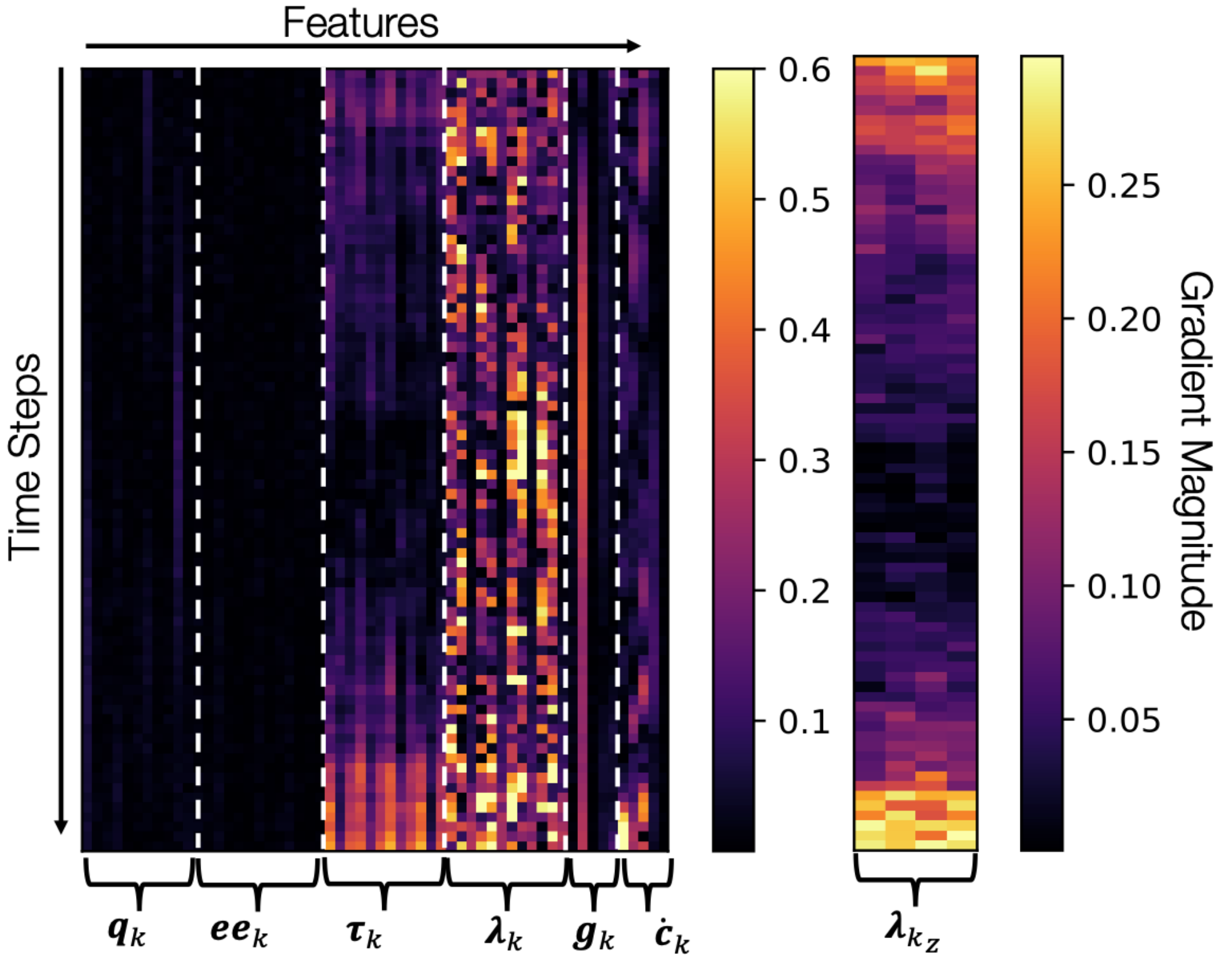}
    \caption{Visualisation of the encoder's receptive field with respect to the gait phase. The encoder input is reshaped such that robot state quantities such as joint torques are in the same column. 
    We also show the contact forces normal to the ground plane $\mathbf{\lambda}_{zk}$. 
    The lightest areas in these two sub-figures are of highest gradient meaning that the encoder focuses on these parts in order to infer the gait phase.}
    \label{fig:saliency_map}
\end{figure}

\begin{figure*}[t]
    \centering
    \includegraphics[width=1.0\linewidth]{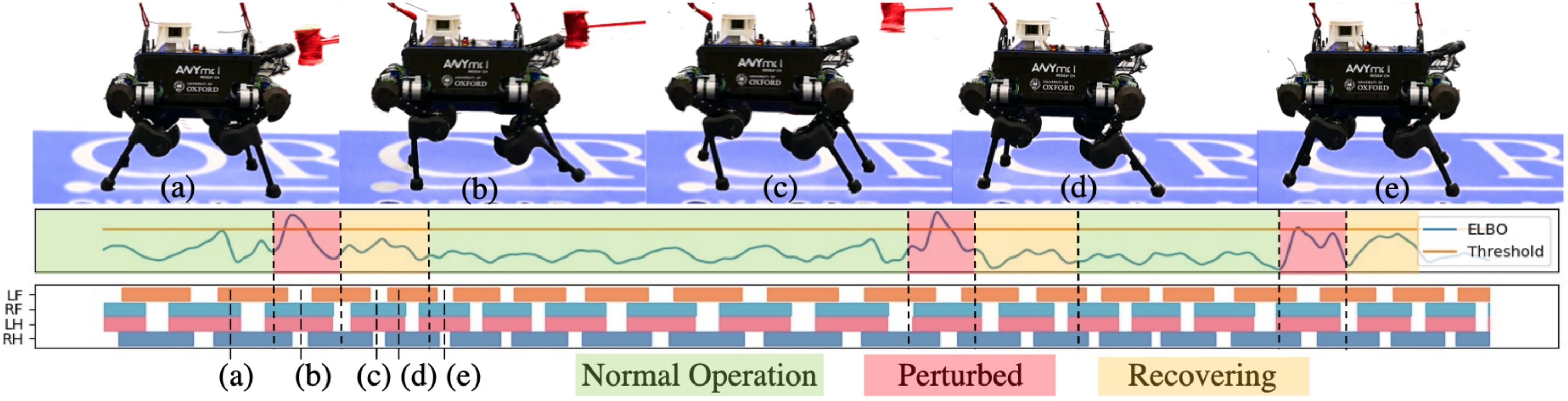}
    \caption{Push recovery following a shove to the base. A disturbance to the robot causes the ELBO of the form in Eq.~\ref{eq:loss_vae} to rise above a predetermined threshold (red areas). This identifies a disturbance and triggers an increase in cadence from \SI{250}{\milli\second} to \SI{125}{\milli\second} as a rudimentary response. This reduction in the swing duration is visible in the contact schedule.
    To appreciate the results fully, please see the following video \url{https://youtu.be/GT2WLh2Ackc}.}
    \label{fig:push_recovery}
\end{figure*}

\subsection{Analysing the encoder's receptive field}\label{section:encoders_receptive_field}
It is crucial that the VAE is able to infer the robot's gait phase from raw sensor input to permit successful latent-space planning.
This poses the question which parts of the encoder input are used to infer this.

Fig.~\ref{fig:saliency_map} depicts a saliency map where a bright colour denotes areas of high gradient.
These are areas where the VAE's encoder focuses to infer the gait phase.
The left map in Fig.~\ref{fig:saliency_map} shows the receptive field of the encoder over the entire input $\rmX_k$.
Here, the input $\rmX_k$ is reshaped such that each row is the robot state $\rvx_k$, see Eq.~\ref{eq:encoder_input}.
The columns represent specific robot quantities which we have highlighted in Fig.~\ref{fig:saliency_map}. 
The areas of high gradient are the contact forces and the most recent joint-torques.
The left part of Fig.~\ref{fig:saliency_map} are the contact forces normal to the ground plane (i.e. the z-direction).
This plot reveals that the most recent and earliest contact forces are utilised to infer the gait phase.
High values of contact force correlate well with the contact state.
The joint torques are related to the contact forces through the robot dynamics' equation~\cite{floating_base_invD}.
Lastly, the base-velocity has a high gradient value.
This quantity is utilised in inferring the robot's momentum.

\subsection{Varying locomotion parameters online} \label{section:sliders}

We leverage the disentangled latent-space to smoothly transition between gait parameters whilst the robot is walking.
Crucially, cadence, stance duration, and footstep height can be varied during any phase of the gait by modulating the drive signal's parameters (see Sec.~\ref{section:drive-signal}).
This results in operating modes which vary from those seen during training.
Examples of walking motion using the VAE-planner on the real robot are shown in Fig.~\ref{fig:cadence-comparison}.

\textbf{Swing Duration:} 
The swing duration is varied over a large operating window on the ANYmal robot.
This begins with a swing time-period starting at \SI{312.5}{\milli\second}, and is smoothly varied until the swing duration reaches \SI{125}{\milli\second}, \revision{more than doubling the step rate}.
In parallel, we alter the robot's heading, demonstrating the independence of the action and the latent-space dynamics.
Specifically, the top row of Fig.~\ref{fig:cadence-comparison} shows the nominal swing duration of \SI{312.5}{\milli\second} as the robot turns clockwise, tracking a constant angular velocity command.
Following this, we demand a slightly faster swing \SI{188}{\milli\second} and a constant angular velocity anti-clockwise, before transitioning to the fastest swing (\SI{125}{\milli\second}) in the third row.
Here, the coloured contact schedule captures the changes in swing duration as it occurs in real-time.

\begin{figure}
    \centering
    \includegraphics[width=1.0\linewidth]{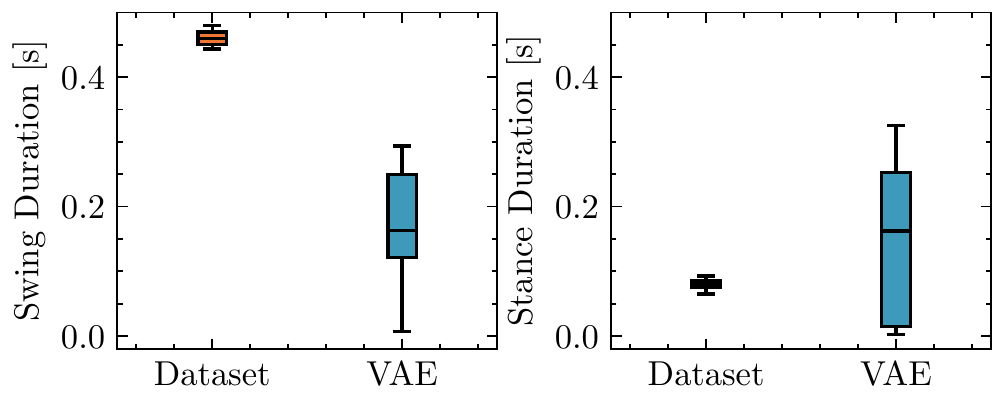}
    \caption{
    Comparison of the distributions of swing and full-stance durations between the dataset used for training and the motions executed on the real robot with the VAE and different drive signal parameters during the test presented in Sec.~\ref{section:sliders}.}
    \label{fig:swing_stance}
    \vspace{-0.25cm}
\end{figure}

\textbf{Stance Duration:} Following a successful reduction in cadence, the stance duration is reduced and the robot transitions into a trot with negligible full-stance phase: $\epsilon_k=0$.
Trot gaits with little to no full-support phase are particularly challenging manoeuvres for the system in general, as there is reduced control authority to correct for accrued base pose error.
During the swing phase of this gait style, only feet across the diagonal are in contact resulting in a line contact limiting the robot's ability to steady its base.
The transition to a negligible full-stance phase is captured in the third row, where the coloured stance duration reduces in length.
\revision{Note that the full-stance duration does not reduce to zero seconds. The mean lowest stance-duration is found to be \SI{19.4}{\milli\second}.  We can see this in Fig.~\ref{fig:swing_stance}, which shows the spread of achievable swing and stance durations compared to the those in the dataset.
The right sub-plot shows the stance durations, where we see that the bottom whisker of plot does not quite touch zero.
The reason for this is that the full-support region in the latent-space must be traversed. This part of latent space is highlighted in purple in Fig~\ref{fig:latent-space}. It is possible to introduce a non-smooth drive-signal which jumps this region, but this discontinuous drive-signal would introduce a large acceleration into the resulting locomotion trajectory. This might not be trackable by the WBC. Instead we recommend utilising a smooth and continuous drive-signal.}

\textbf{Footstep Height:} We vary the amplitude of the drive signal smoothly to zero as is seen in the bottom row (Fig.~\ref{fig:cadence-comparison}): %
The footstep height reduces to zero as the white-space in the contact schedule disappears and the robot remains standing.
Beyond versatility, e.g. to increase swing heights to overcome irregular ground height, this capability further enables a safe, smooth and natural transition into and out of the VAE control mode (i.e. to start and come to a halt).

\begin{figure}
    \centering
    \includegraphics[width=1.0\linewidth]{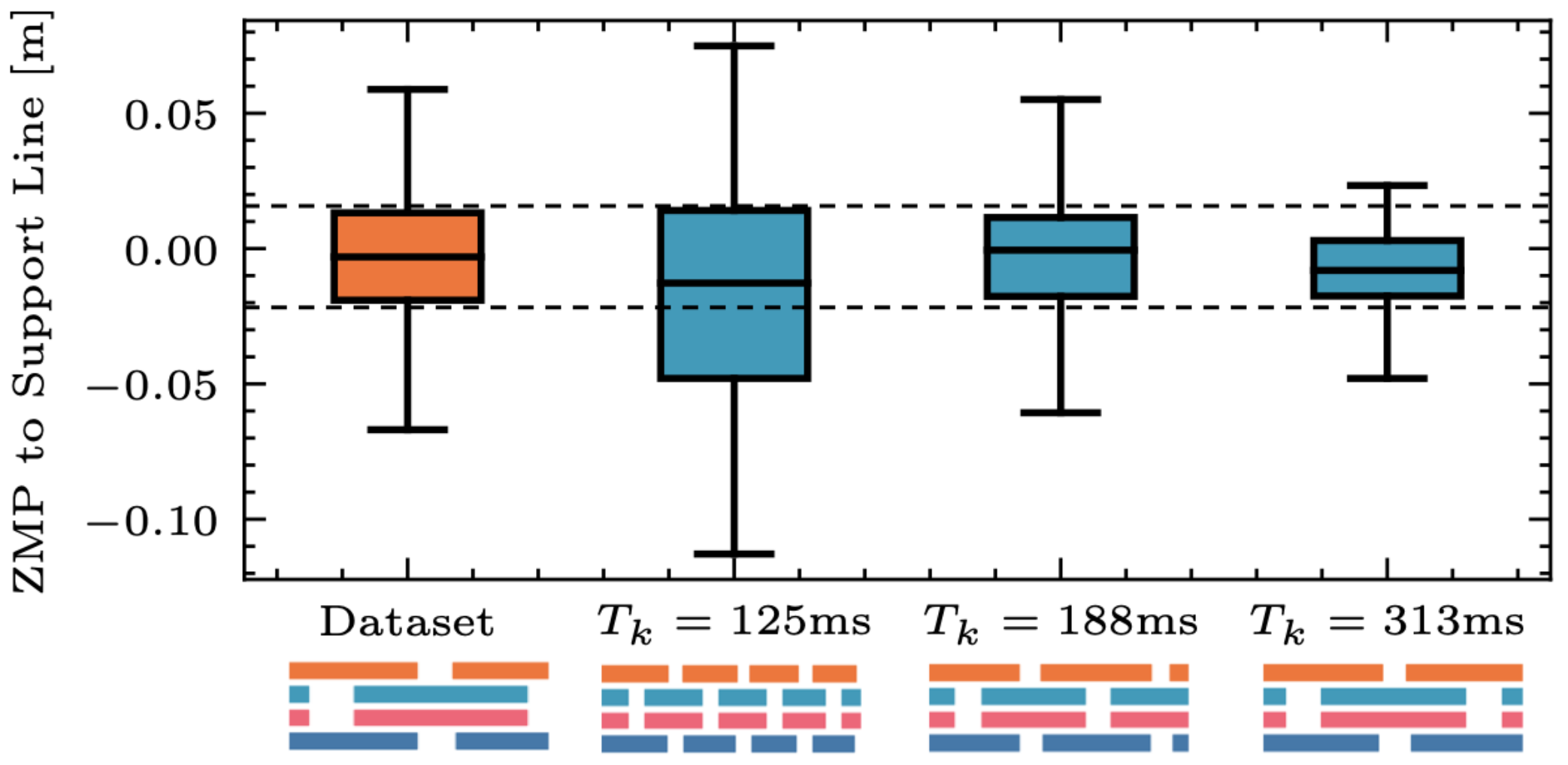}
    \caption{Distribution of the signed distance between the ZMP and the support line during the 2-stance phases (i.e., one pair of legs was in swing). In orange, the distribution with upper and lower bounds (dashed lines) from the dataset 
    which uses a swing duration of $T_k = \SI{0.5}{\second}$. In blue, the distribution of the VAE trajectories shown over a range of modes deployed on the real robot.
    }
    \label{fig:ZMP_comp}
    \vspace{-0.25cm}
\end{figure}

\begin{figure*}[t]
    \centering
    \includegraphics[width=1.0\linewidth]{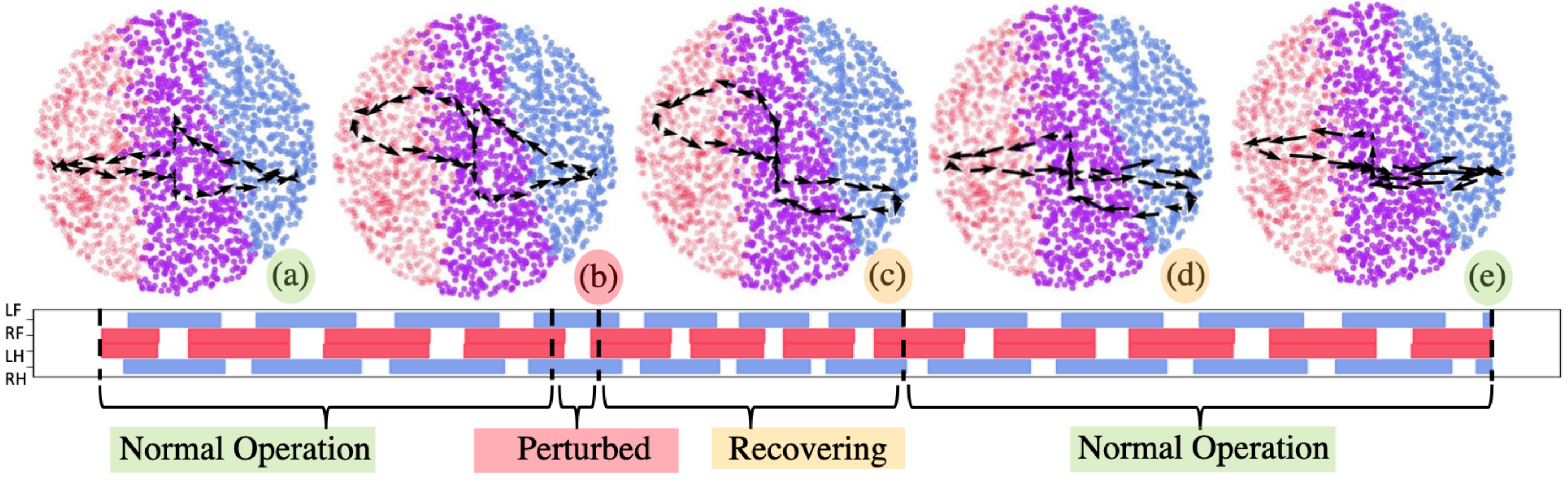}
    \caption{We show the latent-space trajectory in a series of sub-figures as the robot is walking normally, is perturbed, and recovers.
    The latent-space trajectory (in black) is plotted on the latent-space colour-coded to show the stances of the trot-gait.
    In normal operation (sub figures (a) and (e)) the latent-space path is a figure of eight with triangular lobes during the robot's footswing (blue and red areas). As the robot is perturbed (sub figure (b)), the lobes are far more elongated. 
    Crucially, this elongated trajectory reverts back to the nominal trajectory as the robot recovers.
    The accompanying footstep schedule is plotted along side.
    }
    \label{fig:perturbed-latent-space}
    \vspace{-0.25cm}
\end{figure*}

\subsection{\revision{ZMP of planned motion as a measure of dynamic feasibility}} \label{section:evaluating_dynamic_feasibility}

On the robot, the trajectories from the VAE-planner are sent directly to the WBC unaltered.
Here, we compare dynamic feasibility of the trajectories shown in Sec.~\ref{section:sliders} after the experiment in order to compare to the training dataset's distribution generated using DG.
The position of the ZMP relative to the support line is utilised as a metric for this comparison.
The optimal trajectory will have the ZMP lie on the support line.

The results of this are summarised in Fig.~\ref{fig:ZMP_comp}.
The distribution the ZMP positions is similar for both DG (dataset) and a range of the VAE-planner's operating modes. 
Crucially, this remains true despite the maximum swing duration generated by the VAE-planner being up to 3.2 times faster than in the training data.
We conclude that the representation is good enough to generalise to the robot's dynamics shown here.
Empirically, we have further been able to steer the robot with arbitrary and fast changing input actions for x, y, and yaw rates, issued from a remote control while being able to interpolate the gait style.

\begin{figure}
    \centering
    \includegraphics[width=0.85\linewidth]{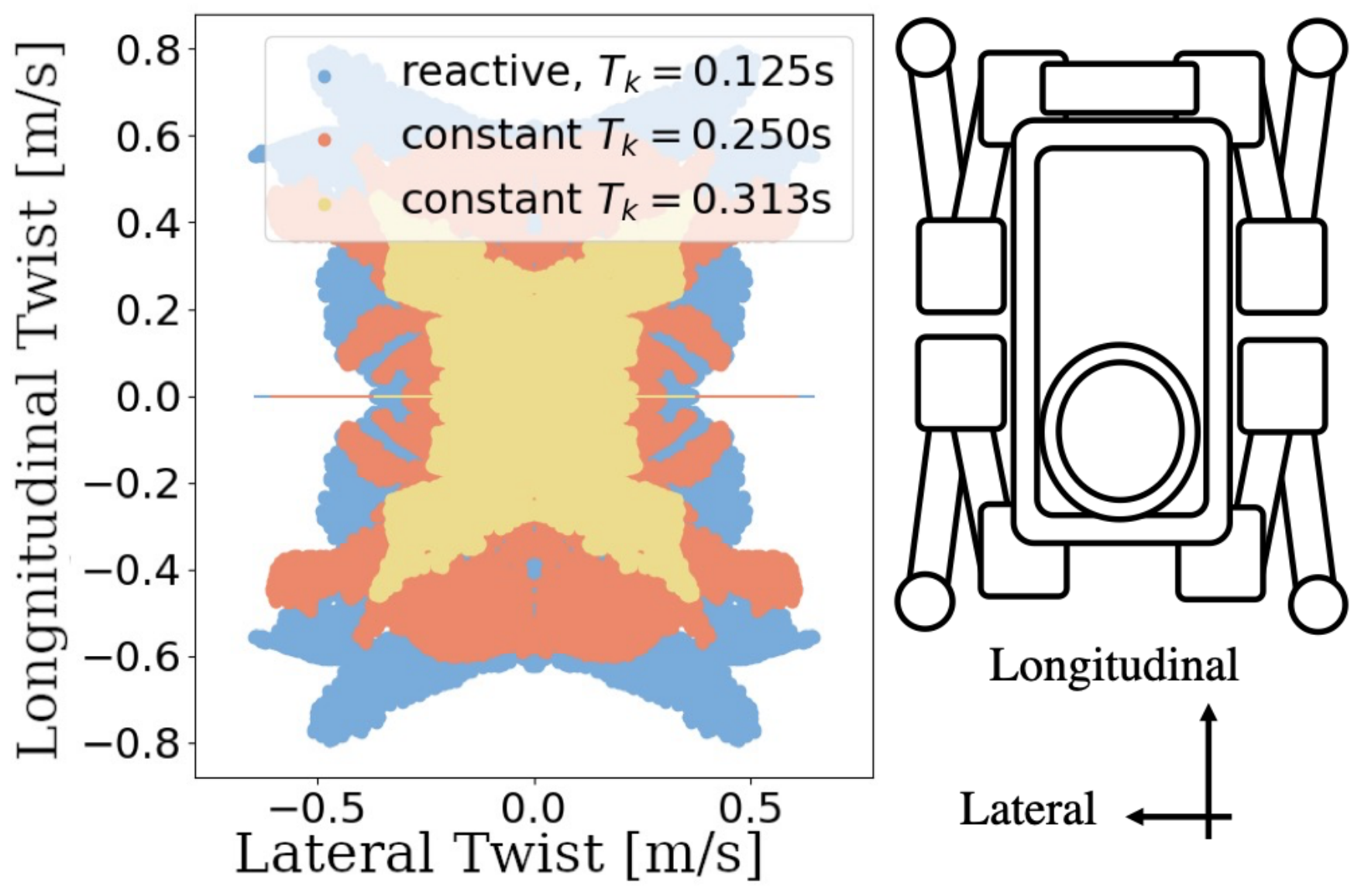}
    \caption{The range of rejectable base-velocity disturbances increases when automatically modulating the cadence.
    This plot is symmetrical for ease of view.
    The range of swing durations plotted is \SI{313}{\milli\second}, \SI{250}{\milli\second}, and \SI{125}{\milli\second}.}
    \label{fig:base_twist_rejection}
    \vspace{-0.25cm}
\end{figure}

\subsection{Generalisation to different \revision{trot-}gait styles} \label{section:generalisation_gait}
The VAE is able to produce a far broader range of different trot styles than those in the dataset.
The ranges of swing and stance durations produced by the VAE in Sec.~\ref{section:sliders} and those in the dataset are summarised in Fig.~\ref{fig:swing_stance}.
As discussed previously, the dataset is constructed utilising a pre-determined and constant swing duration of \SI{500}{\milli\second}, with stance duration of \SI{75}{\milli \second}.
The VAE-planner's swing duration varies between \SI{312.5}{\milli \second} and \SI{125}{\milli\second} and its stance ranges between \SI{325}{\milli\second} to negligible duration. 
This demonstrates that the VAE-planner is able to generalise to a variety of trot styles by navigating the structure within the disentangled latent-space, even when trained using a very restrictive training set.

\subsection{Disturbance detection and recovery}\label{section:disturbance_rejetion}

\revision{Our approach is able to detect and mitigate disturbances applied to the robot using a heuristic policy which monitors the ELBO loss whilst the robot is walking.}
\revision{The robot is pushed with a broom resulting in a large spike in the ELBO loss.}
\revision{A spike above a certain threshold is characterised as a disturbance.}
For our push experiments, we choose the ELBO threshold value to be ${11.0}$.
We find this value by walking the robot around the lab using the VAE-planner and choosing a value above what is seen during normal operation.

The VAE-planner \revision{in nominal conditions} is able to reject a wide range of impulses applied to the robot’s base. 
However, this operating window is enlarged by increasing the robot’s cadence as soon as a disturbance is detected.
This \emph{reactive} VAE-planner halves the robot's swing duration from a nominal \SI{250}{\milli\second} to \SI{125}{\milli\second} once the ELBO threshold is surpassed.

We show the VAE-planner detecting and reacting to the push broom disturbance in Fig.~\ref{fig:push_recovery}.
The top row shows the robot being pushed violently before recovering in three to four steps.
At point (b) the ELBO spikes above the threshold indicating the disturbance.
Throughout the next \SI{1.5}{\second}, the cadence is halved.
This is visible in the contact schedule---the white space reduces in width, see point (c).
At points (d) and (e) the robot has fully recovered.
The ELBO and contact schedule plots continue on and the robot is pushed twice more in this figure.

\begin{figure*}[t]
    \centering
    \includegraphics[width=1.0\linewidth]{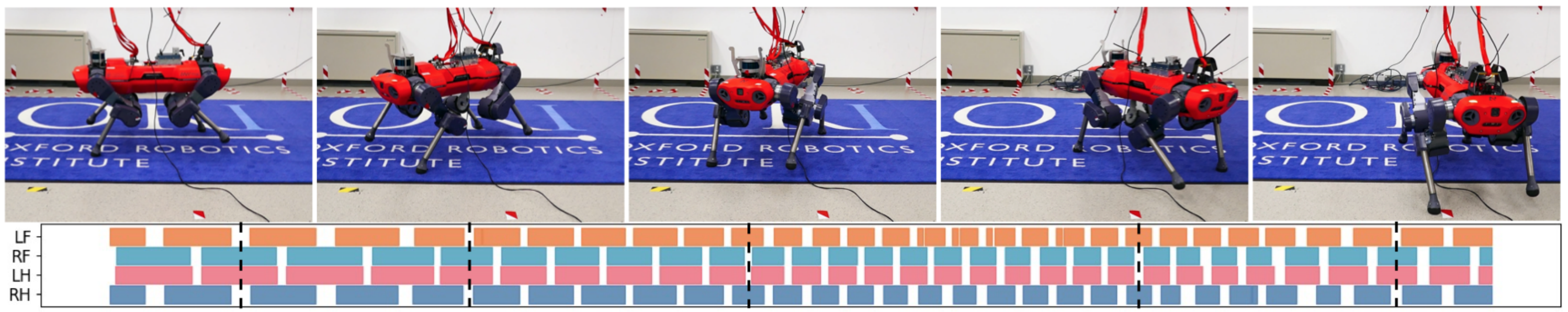}
    \caption{We deploy our VAE-planner on a different robot: ANYmal C. The VAE used here is trained using data from a simulated ANYmal B robot.
    ANYmal C is significantly heavier than ANYmal B (\SI{50}{\kilo\gram} to \SI{35}{\kilo\gram}) with twice as much maximum torque, but has actuators with lower bandwidth.
    With the use of the WBC, the VAE-planner is able to control ANYmal C effectively.
    The user control over the robot's base twist using the action and the gait parameters are varied as on ANYmal B using the drive-signal parameters.
    The variation in gait parameters is captured by the swing and stance durations in the contact schedule below the images of the robot. Please view the ANYmal C locomotion in our video found at \url{https://youtu.be/GT2WLh2Ackc}.}
    \label{fig:coyote}
\end{figure*}

A comparison between the reactive VAE-planner and a constant cadence version is drawn in Fig.~\ref{fig:base_twist_rejection}.
The constant cadence VAE-planner uses swing durations of \SI{313}{\milli\second} and \SI{250}{\milli\second}, whilst the reactive version halves the robot's cadence to \SI{125}{\milli\second} as described.
Here, we see that the range of rejectable push disturbances increases as the swing duration decreases.

Finally, we illustrate what happens to the latent-space trajectory during and after the disturbance.
The nominal latent-space trajectory forms a figure of eight and is typified by triangular lobes. 
This is visible in Fig.~\ref{fig:perturbed-latent-space} sub-figures (a) and (e). 
When the robot is perturbed these lobes deform and elongate e.g. in sub-figures (b) and (c).
The latent-space trajectory converges back to nominal causing the robot to recover as in sub-figures (c) and (d).
Finally, the robot has fully recovered and the latent-space trajectory return to nominal (sub-figure (e)).

\subsection{Transfer to ANYmal C}\label{section:anymal_c}
We successfully deploy the VAE-planner trained using simulated ANYmal B data on ANYmal C.
Please note that the VAE is not retrained.
The differences between the two robots are significant as ANYmal C's joint torque limits are \SI{80.0}{\newton\meter} instead of \SI{40.0}{\newton\meter}, and ANYmal C's total mass is \SI{50}{\kilo\gram} instead of \SI{35}{\kilo\gram}.
Also note that the WBC's internal model for both kinematics and dynamics are updated when the VAE-planner is deployed on ANYmal C.
However, the joint trajectory output from the VAE-planner is the input to the WBC meaning that joint trajectories from the VAE-planner are suitable for ANYmal C.

Despite not retraining, the VAE-planner is able to command ANYmal C's heading and control the robot's gait parameters.
In essence, the experiment in Sec.~\ref{section:sliders} is repeated with the same VAE-planner, but deployed on ANYmal C.
Fig.~\ref{fig:coyote} shows ANYmal C with the VAE-planner deployed onboard, along with the contact schedule.
The latter shows the variation in both swing and stance duration as the robot walks.

\textbf{Comparing ANYmal B and C using ELBO:}
The ELBO can be used to further compare the differences between encoding the raw ANYmal B and C data. 
As mentioned in Sec.~\ref{section:exp_design_deployment_anymal_c}, key parameters such as joint torque limits are standardised before encoding.
Analysis of the ELBO is split into comparison of the KL divergence and mean-squared reconstruction error between ANYmal B and ANYmal C.

The KL divergence term is particularly affected by latency in the input.
For example, if the states in the encoder input are not sampled exactly at the encoder frequency, the KL divergence term will increase.
The distributions of these values are quite different between each robot.
The KL-divergence values for ANYmal B show a steady mean drift upwards which is negligible for ANYmal C. 
The median value for ANYmal C is also noticeably lower than for ANYmal B.

As described in Sec.~\ref{section:exp_design_deployment_anymal_c}, we compare the KL-divergence values from both robots together using a Mann-Whitney U-test~\cite{mann1947}.
As a reminder, we record the KL-divergence for both robots and analyse sets of size $2.5\times 10^4$.
The resulting U-statistic is $1.82\times 10^9$ and the p-value is $p<0.001$.
Therefore, the null-hypothesis is rejected and the difference in the median values of the KL-divergence from both robots is statistically significant.

The reason for ANYmal C's lower median value is down to the relative performance of the onboard computers on the two ANYmals.
ANYmal C has a much faster CPU meaning that the VAE-planner's control loop is comfortably under the \SI{2.5}{\milli\second} required for real-time control.
The result of which is the encoder's input is sampled at the desired encoder frequency. 
In contrast, on average the control loop on ANYmal B violates the real-time control law roughly \SI{20}{\percent} of the time during the experiment run in Fig.~\ref{fig:swing_stance}.
This introduces latency into the control loop which causes the increase in KL-divergence.

The reconstruction error for ANYmal B is, unsurprisingly, lower than for ANYmal C. 
The median value for ANYmal B reconstruction error is $5.74\times 10^{-3}$, whilst for ANYmal C it is $1.58\times 10^{-2}$. 
These median values are compared using the Mann-Whitney U-test conducted over two sets containing $2.5\times 10^4$ values.
The U-statistic is $8.26\times 10^7$ resulting in a p-value of $p<0.001$.
The results in the rejection of the null-hypothesis, meaning that the two distributions are statistically different.
Since the VAE-planner is trained using data from the ANYmal B, a lower reconstruction error is expected.

\section{Future work}\label{section:future_work}
\rev{The most significant limitation of the current system is that, despite the VAE-planner being able to interpolate between a broad range of dynamic manoeuvres, the resulting locomotion gaits are constrained to follow the trot contact sequence and operate on flat ground. This stems from the decision to only train the model using demonstrations of the trot gait with fixed parameters and on flat ground. These decisions were made so that we could inspect and interpret the resulting latent-space thoroughly in this restrictive domain. 

To generate different contact schedules other than trot, it is necessary to train a new VAE with different gait trajectories. These different contact schedules should similarly become embedded in latent space. This provides an opportunity to train the VAE with even more dynamic gaits such as bound as well as asymmetrical gaits. 

Another limitation is that the current VAE-planner has not been exposed to the trajectories required in order to traverse uneven terrain. Therefore, we wish to train the model on a variety of different locomotion gaits generated when the quadruped walks in unstructured environments. The resulting latent-space will be analysed and the VAE deployed as a planner so that the quadruped can operate in these environments. This will discover firstly how well the VAE-planner reacts to external disturbances resulting from uneven terrain, and secondly the range of motion the VAE-planner can produce in response to the unstructured terrain.

}

\section{Conclusion}\label{section:conclusion}

In this paper, we present a robust and flexible approach for locomotion planning from the perspective of traversing a structured latent-space.
This is achieved utilising a deep generative model to capture relevant structure from locomotion data and enables the detection and mitigation of disturbances.
The latent space is disentangled to a degree that key salient locomotion features are automatically discovered from a single style of trot gait.
An investigation of this latent space reveals a two-dimensional representation which encapsulates the underlying dynamics of the system. 
This disentanglement is exploited using a drive signal with which dynamically consistent locomotion is generated. 
Crucially, the amplitude and phase of the drive signal directly control the gait characteristics, namely the cadence, swing height, and full-support duration.
Once deployed, the ease with which modulation of the drive signal gives rise to seamless interpolation between gait parameters is demonstrated.
Despite generalising to remarkably distinct trot styles compared to the training distribution, the entire range of VAE trajectories remains dynamically consistent.
Additionally, utilising a generative model affords the ability to characterise disturbances as out of the distribution seen during training.
Though the VAE-planner is able to reject a broad range of impulses applied to the robot's base, this window is broadened by increasing the cadence as soon as the disturbance is detected.
Finally, we show that the approach readily transfers to a kinematically similar but dynamically different platform without needing to be retrained.
This helps to showcase the VAE's ability transfer to new unseen domains.  

\section*{Acknowledgements}
The authors would like to acknowledge the use of the SCAN facility, and thank Oliver Groth for useful discussions.

\bibliographystyle{IEEEtran}
\bibliography{references}

\begin{IEEEbiography}[{\includegraphics[width=1in,height=1.25in,clip,keepaspectratio]{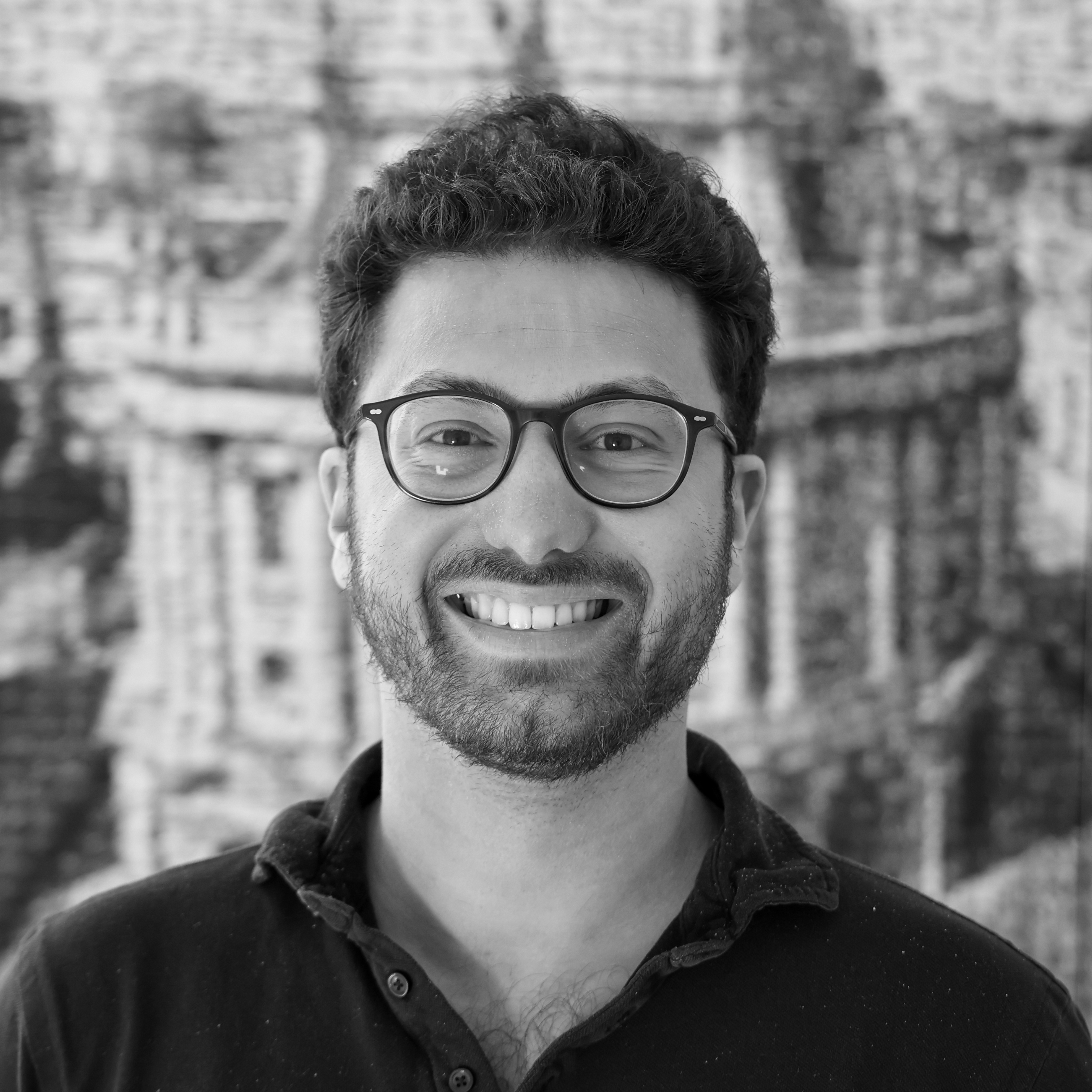}}]%
    {Alexander L. Mitchell} received his M.Eng. from the University of Oxford in 2018. Subsequently, he enrolled at the University of Oxford as a Doctoral candidate. He is currently co-supervised by Professor Ingmar Posner and Dr Ioannis Havoutis at the Oxford Robotics Institute. 
    Alexander's research interests include optimal control and learning-based methods for path planning and control of legged robots.
\end{IEEEbiography}

\begin{IEEEbiography}[{\includegraphics[width=1in,height=1.25in,clip,keepaspectratio]{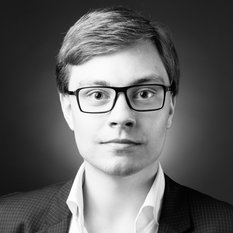}}]%
    {Wolfgang Merkt} received the B.Eng.(Hns) degree in mechanical engineering with management and the M.Sc.(R) and Ph.D. degrees in robotics and autonomous systems from the University of Edinburgh, Edinburgh, U.K., in 2014, 2015 and 2019, respectively.
    
    He is currently a Senior Researcher at the Oxford Robotics Institute, University of Oxford with I. Havoutis.
    During his Ph.D., he worked on trajectory optimization and warm starting optimal control for high-dimensional systems and humanoid robots under the supervision of S. Vijayakumar.
    His research interests include fast learning- and optimization-based methods for planning and control, loco-manipulation, and legged robots.
\end{IEEEbiography}

\begin{IEEEbiography}[{\includegraphics[width=1in,height=1.25in,clip,keepaspectratio]{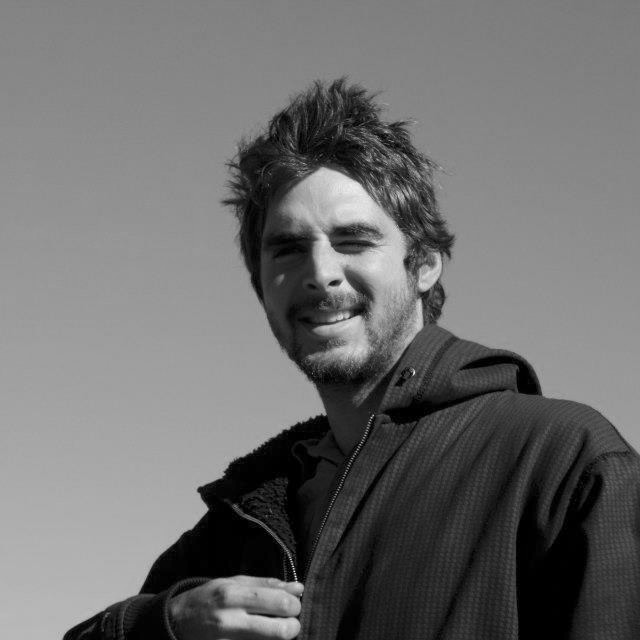}}]%
    {Mathieu Geisert} received his M. Eng. in Aerospace from Institut Supérieur de l’Aéronautique et de l’Espace SUPAERO (Toulouse, France) in 2013. He then joined the humanoid robotics team Gepetto at LAAS-CNRS (Toulouse, France) where he worked for 5 years in different positions and received his PhD in 2018. From 2018 to 2021, he worked on quadruped robots as a post doctoral researcher in the Dynamic Robot Systems (DRS) group (Oxford Robotics Institute - University of Oxford). From 2021, he has been working as a controls research engineer at Arrival.
    His research focuses on Control and Learning for Legged Robots Locomotion.
\end{IEEEbiography}

\begin{IEEEbiography}[{\includegraphics[width=1in,height=1.25in,clip,keepaspectratio]{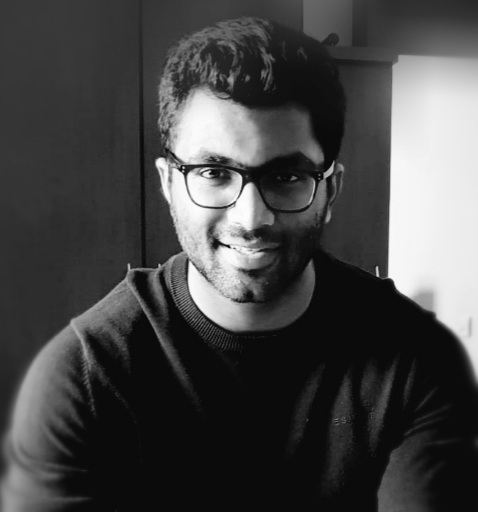}}]%
    {Siddhant Gangapurwala} completed his Bachelor of Engineering (B.E.) in Electronics from the University of Mumbai in 2016. In 2017, he joined the AIMS program as a doctoral candidate at the University of Oxford. The following year, Siddhant joined the Dynamic Robot Systems (DRS) group of the Oxford Robotics Institute (ORI) to pursue his DPhil degree with focus on machine learning and optimal control based approaches for robotic locomotion over uneven terrain. He continues to work at DRS as a post-doctoral researcher on locomotion, manipulation and loco-manipulation.
\end{IEEEbiography}

\begin{IEEEbiography}[{\includegraphics[width=1in,height=1.25in,clip,keepaspectratio]{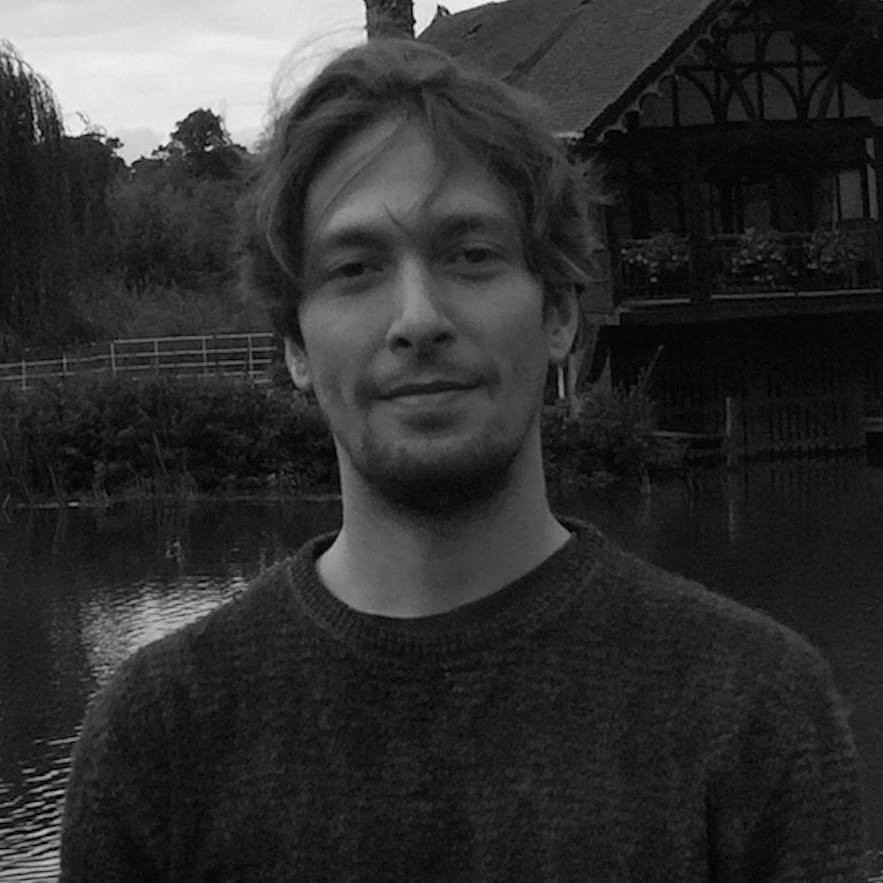}}]%
    {Oiwi Parker Jones} received a D.Phil. in natural language processing at the University of Oxford, Oxford, U.K. From there he went on to do postdoctoral work in language neuroscience, computational modelling, and the development of neuroimaging methods for large datasets at University College London and at the University of Oxford. He is currently a senior postdoc in the Applied AI Lab (A2I) at the Oxford Robotics Institute, honorary fellow in the Nuffield Department of Clinical Neurosciences, and Hugh Price Fellow in Computer Science at Jesus College, University of Oxford.
\end{IEEEbiography}

\begin{IEEEbiography}[{\includegraphics[width=1in,height=1.25in,clip,keepaspectratio]{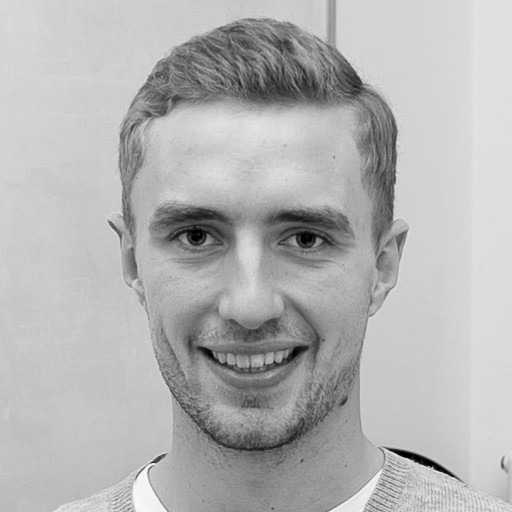}}]%
    {Martin Engelcke} received the M.Eng. and D.Phil. (Ph.D.) from the University of Oxford, Oxford, UK. He conducted his doctoral and subsequent postdoctoral research between 2015 and 2021 at the Oxford Robotics Institute under the supervision of Prof. Ingmar Posner. He is now a research scientist at DeepMind. His research interests include generative models, representation learning, and reinforcement learning.
\end{IEEEbiography}

\begin{IEEEbiography}[{\includegraphics[width=1in,height=1.25in,clip,keepaspectratio]{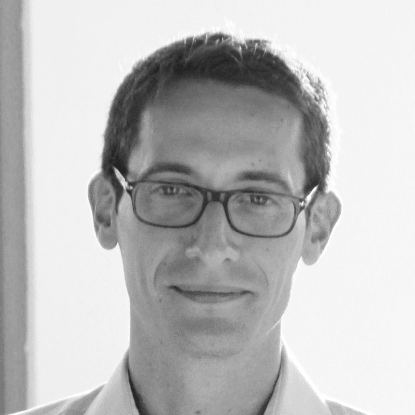}}]%
    {Ioannis Havoutis} received the M.Sc. in artificial intelligence and Ph.D. in Informatics degrees from the University of Edinburgh, Edinburgh, U.K.
    
    He is a Lecturer in Robotics at the University of Oxford.
    He is part of the Oxford Robotics Institute and a co-lead of the Dynamic Robot Systems group.
    His focus is on approaches for dynamic whole-body motion planning and control for legged robots in challenging domains.
    From 2015 to 2017, he was a postdoc at the Robot Learning and Interaction Group, at the Idiap Research Institute.
    Previously, from 2011 to 2015, he was a senior postdoc at the Dynamic Legged System lab the Istituto Italiano di Tecnologia.
    He holds a Ph.D. and M.Sc. from the University of Edinburgh.
\end{IEEEbiography}

\begin{IEEEbiography}[{\includegraphics[width=1in,height=1.25in,clip,keepaspectratio]{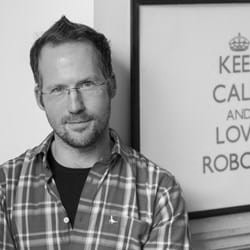}}]%
    {Ingmar Posner} is Professor of Engineering Science at Oxford University, where he leads the Applied Artificial Intelligence Lab. His research aims to enable machines to robustly act and interact in the real world - for, with, and alongside humans. Ingmar’s track record in machine perception and decision-making includes seminal work on large-scale learning from demonstration, representation learning for scene understanding and prediction as well as 3D object detection and machine introspection. Currently his research interests revolve around the use of structured latent spaces for robot perception, planning and control.
    
    Ingmar received an M.Eng. degree in Electronic Systems Engineering from Aston University, Birmingham, U.K., and a D.Phil. degree in bioacoustics from the University of Oxford. He is a founding Director of the Oxford Robotics Institute and, in 2014, co-founded Oxbotica, multi-award winning provider of mobile autonomy software solutions.
\end{IEEEbiography}

\end{document}